\documentclass[lettersize,journal]{IEEEtran}
\usepackage{amsmath,amsfonts}
\usepackage{array}
\usepackage{textcomp}
\usepackage{stfloats}
\usepackage{url}
\usepackage{verbatim}
\usepackage{graphicx}
\usepackage{cite}
\usepackage{marvosym}
\usepackage{epsfig}
\usepackage{graphicx}
\usepackage{textcomp}
\usepackage{amsmath}
\usepackage{mathtools}
\usepackage[linesnumbered,ruled,vlined]{algorithm2e}

\SetCommentSty{mycommfont}
\usepackage{multirow}
\usepackage{array}
\usepackage{setspace}
\usepackage{caption}
\usepackage{subcaption}
\usepackage{float}
\usepackage[autostyle]{csquotes}

\usepackage[utf8]{inputenc} 
\usepackage[T1]{fontenc}    
\usepackage{hyperref}       
\usepackage{url}            
\usepackage{booktabs}       
\usepackage{amsfonts}       
\usepackage{nicefrac}       
\usepackage{microtype}      
\usepackage{xcolor}         
\usepackage{epsfig}
\usepackage{graphicx}
\usepackage{amsmath}
\usepackage{amssymb}
\usepackage{textcomp}
\usepackage{mathtools}
\usepackage[ruled,vlined]{algorithm2e}
\usepackage[autostyle]{csquotes}
\SetCommentSty{mycommfont}
\usepackage{multirow}
\usepackage{array}
\usepackage{wrapfig}
\usepackage{enumitem}
\setlist{topsep=0pt, leftmargin=*}
\usepackage{setspace}
\usepackage[stable]{footmisc}
\usepackage{caption}
\usepackage{subcaption}
\usepackage{float}

\def\TIBBIMN{TI-BNIM\textsubscript{N}\xspace}
\def\TIBBIMK{TI-BNIM\textsubscript{K}\xspace}

\def\FGBMN{FGNM\textsubscript{N}\xspace}
\def\FGBMK{FGNM\textsubscript{K}\xspace}
\def\inca{Inc-v3\xspace}
\def\incb{Inc-v4\xspace}
\def\res{Res152\xspace}
\def\den{Den161\xspace}
\def\inre{IncRes\xspace}
\def\incaE{Inc-v3\textsubscript{ens3}\xspace}
\def\incbE{Inc-v3\textsubscript{ens4}\xspace}
\def\inreE{IncRes\textsubscript{ens}\xspace}
\newcommand{\eat}[1]{}
\newcolumntype{P}[1]{>{\centering\arraybackslash}p{#1}}

\newcommand{\norm}[1]{\left\lVert#1\right\rVert}
\DeclarePairedDelimiter{\abs}{\lvert}{\rvert}
\DeclareMathOperator*{\argmax}{arg\,max}

\DeclareMathOperator{\clip}{clip}

\DeclareMathOperator{\sign}{sign}

\DeclarePairedDelimiterX\set[1]\lbrace\rbrace{#1}
\DeclarePairedDelimiter{\innerprod}\langle\rangle

\newcommand{\smallsim}{\smallsym{\mathrel}{\sim}}

\newcommand{\vect}[1]{\boldsymbol{#1}}

\SetKwInput{KwInput}{Inputs}                
\SetKwInput{KwOutput}{Outputs}              

\makeatletter
\DeclareRobustCommand\onedot{\futurelet\@let@token\@onedot}
\def\@onedot{\ifx\@let@token.\else.\null\fi\xspace}

\def\eg{\emph{e.g}\onedot} 
\def\ie{\emph{i.e}\onedot} 
 
\def\etc{\emph{etc}\onedot} 
\def\wrt{w.r.t\onedot} 
\def\etal{\emph{et al}\onedot}

\newcommand{\smallsym}[2]{#1{\mathpalette\make@small@sym{#2}}}
\newcommand{\make@small@sym}[2]{%
  \vcenter{\hbox{$\m@th\downgrade@style#1#2$}}%
}
\newcommand{\downgrade@style}[1]{%
  \ifx#1\displaystyle\scriptstyle\else
    \ifx#1\textstyle\scriptstyle\else
      \scriptscriptstyle
  \fi\fi
}

\newcommand{\subalign}[1]{%
  \vcenter{%
    \Let@ \restore@math@cr \default@tag
    \baselineskip\fontdimen10 \scriptfont\tw@
    \advance\baselineskip\fontdimen12 \scriptfont\tw@
    \lineskip\thr@@\fontdimen8 \scriptfont\thr@@
    \lineskiplimit\lineskip
    \ialign{\hfil$\m@th\scriptstyle##$&$\m@th\scriptstyle{}##$\hfil\crcr
      #1\crcr
    }%
  }%
}

\makeatother

\begin{document}

\title{Fast Gradient Non-sign Methods}
\author{Yaya Cheng, Jingkuan Song, Xiaosu Zhu, Qilong Zhang, Lianli Gao, Heng Tao Shen~\IEEEmembership{Fellow, IEEE}
\thanks{}
\thanks{Y.~Cheng, J.~Song, X.~Zhu, Q.~Zhang, L.~Gao, and H.~T.~Shen are with Center for Future Media and School of Computer Science and Engineering, University of Electronic Science and Technology of China, Chengdu, 611731, China~(Email: \url{yaya.cheng@hotmail.com})}
\thanks{Manuscript received XX.}}

\markboth{Journal of \LaTeX\ Class Files,~Vol.~14, No.~8, August~2021}%
{Shell \MakeLowercase{\textit{et al.}}: A Sample Article Using IEEEtran.cls for IEEE Journals}

\IEEEpubid{0000--0000/00\$00.00~\copyright~2021 IEEE}

\maketitle

\begin{abstract}
Adversarial attacks make their success in \enquote{fooling} DNNs, and among them, gradient-based algorithms become one of the mainstreams. Based on the linearity hypothesis~\cite{fgsm}, under $\ell_\infty$ constraint, $sign$ operation applied to the gradients is a good choice for generating perturbations. However, side-effects from such operation exist since it leads to the bias of direction between real gradients and perturbations. In other words, current methods contain a gap between real gradients and actual noises, which leads to biased and inefficient attacks. Therefore in this paper, based on the Taylor expansion, the bias is analyzed theoretically, and the correction of $\sign$, \ie, Fast Gradient Non-sign Method (FGNM), is further proposed. Notably, FGNM is a general routine that seamlessly replaces the conventional $sign$ operation in gradient-based attacks with negligible extra computational cost. Extensive experiments demonstrate the effectiveness of our methods. Specifically, for untargeted black-box attacks, ours outperform them by \textbf{27.5\%} at most and \textbf{9.5\%} on average. For targeted attacks against defense models, it is \textbf{15.1\%} and \textbf{12.7\%}. Our anonymous code is publicly available: \url{https://github.com/yaya-cheng/FGNM}.
\end{abstract}



\section{Introduction}
\label{Sec.Intro}

\IEEEPARstart{R}{esearchers} have noticed that, by applying human-imperceptible perturbations, some \enquote{special} inputs will lead DNNs to give unreasonable outputs, which facilitates the born of adversarial attacks~\cite{fgsm,ifgsm,mifgsm,difgsm,tifgsm,pifgsm,pofgsm,deepPrior,attack5,nan,love}. Although a series of DNNs have shown their incredible performance in computer vision tasks, their vulnerability is still a severe problem as their explosive development in recent years~\cite{physicalAttack1,physicalAttack2,physicalAttack4,physicalAttack11,physicalAttack12,physicalAttack13,attack_saliency_TIP}. It is vital to learn to generate adversarial examples since it can help to evaluate and improve robustness of DNNs. 

\begin{figure}[t]
    \begin{center}
       \includegraphics[width=1\columnwidth]{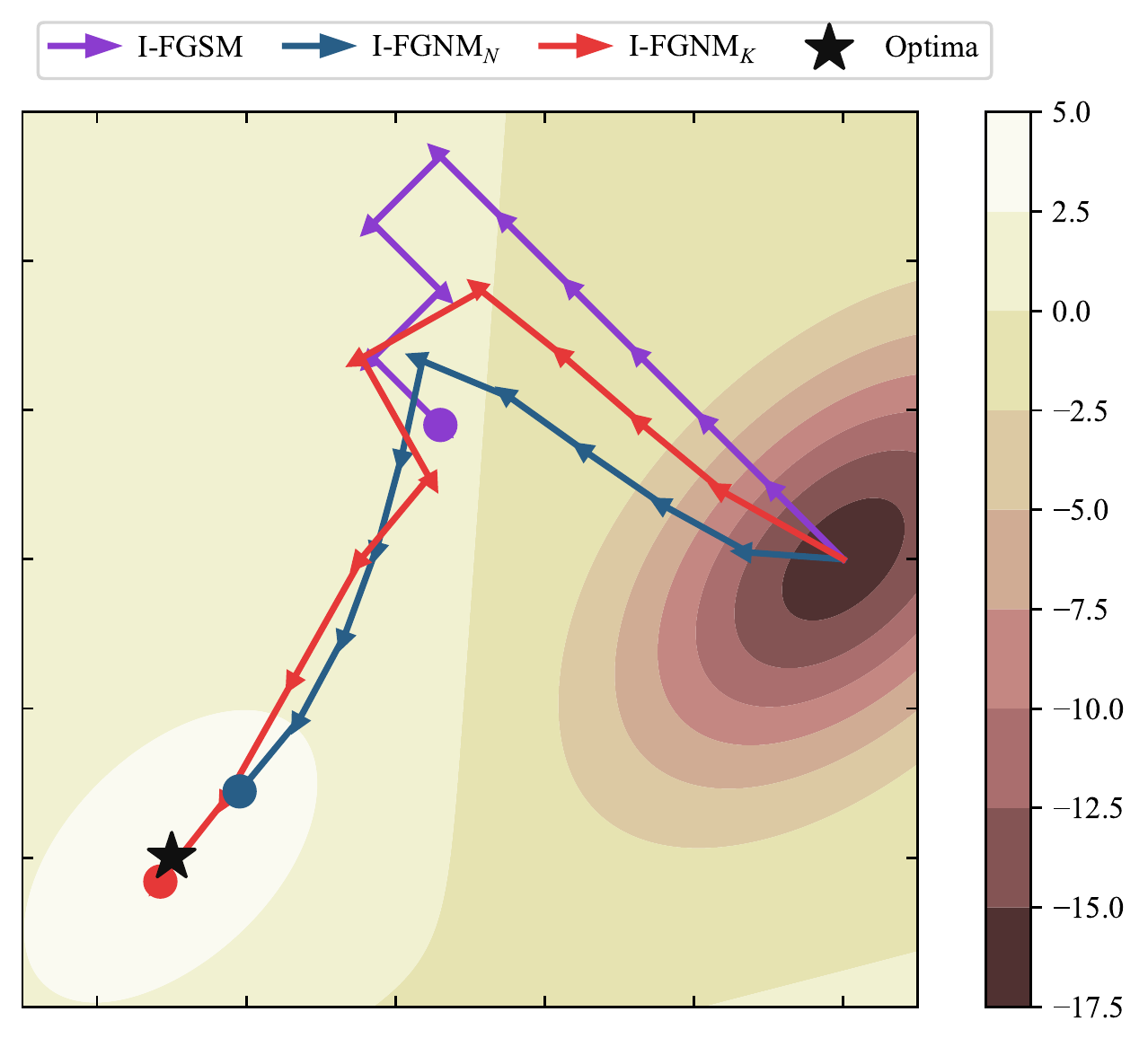}
    \end{center}
   \caption{Visualization on a toy example of climbing mountain with traditional I-FGSM and our I-FGNM variants. Specifically, when adopting I-FGSM, the trajectory is biased due to the limitation of $\sign$. It can not approach the optima and results in oscillation. Our I-\FGBMN holds the identical step size with I-FGSM in each iteration, but produces a more precise trajectory and is close to the optima. The I-\FGBMK considers the trade-off between magnitude and precision in each step and gives the best result.}
    \label{fig:vis}
\end{figure}
\begin{figure}
    \centering
    \begin{subfigure}[b]{0.48\columnwidth}
         \centering
         \includegraphics[width=\textwidth]{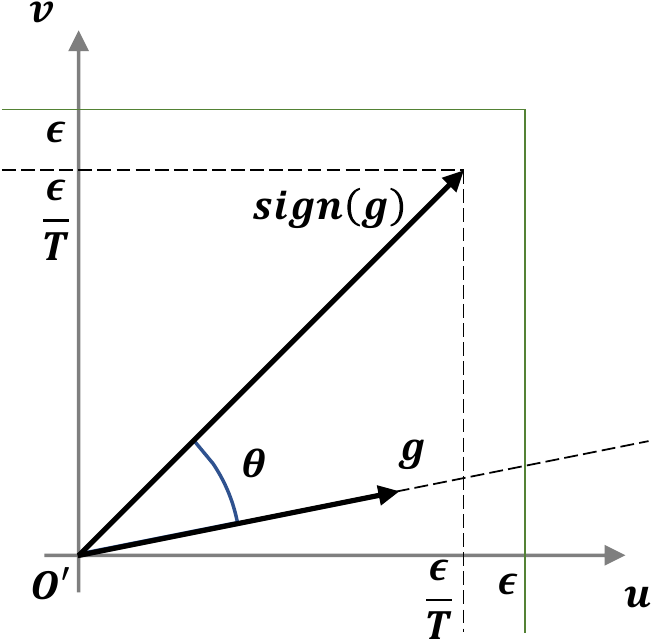}
         \caption{}
    \end{subfigure}
    \hfill
    \begin{subfigure}[b]{0.48\columnwidth}
         \centering
         \includegraphics[width=\textwidth]{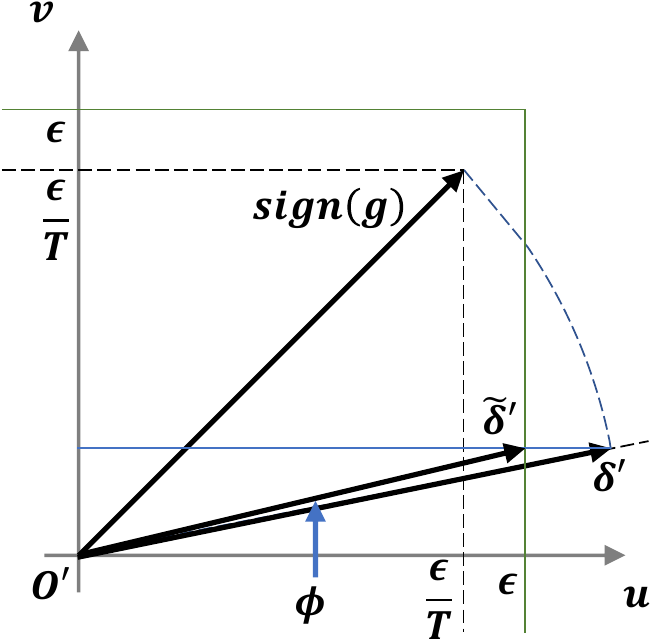}
         \caption{}
    \end{subfigure}
    \caption{Demonstration of the side-effect of $\sign$ operation in the $2D$ Euclidean space in the last step. (a): After $\sign$, there exist a included angle $\angle\theta$ between gradient $\vect{g}$ and $\sign{\left(\vect{g}\right)}$, (b): Our \FGBMN is done by firstly keeping magnitude same as $\sign\left(\vect{g}\right)$, directions same as $\vect{g}$ and then being clipped. Easily verified in the figure, the included angle $\phi$ between $\tilde{\vect{\delta'}}$ and $\vect{g}$ is smaller than $\theta$.}
    \label{fig:hero}
\end{figure}

Several methods have been proposed to craft adversarial examples. In general, adversarial attacks can be grouped into two categories: non-targeted attacks and targeted attacks. For targeted attacks, adversaries should induce victim's models to output specific wrong label. For non-targeted attacks, adversarial example is expected to mislead victim's models to output an arbitrary class except the original one. As a popular branch both in targeted and non-targeted attacks, gradient-based methods focus on how to produce proper perturbations based on gradients, such as one-step attacks~\cite{fgsm,tifgsm} and iterative-step attacks~\cite{ifgsm,mifgsm,difgsm}. By taking the ascending direction of gradient of an objective function $\mathcal{L}$, \eg, cross-entropy, perturbations are obtained and adversarial examples are pulled away from the original decision boundary. To make adversarial examples further away from this decision boundary and reduce computational overhead, most gradient-based methods decide to craft adversarial examples by performing $\sign$ operations according to the linearity hypothesis~\cite{fgsm}. In particular, given a linear model $\mathcal{F}\left(\vect{x}\right) = \vect{w}^{\mathsf{T}}\vect{x}+\vect{b}$, adversarial perturbations $\vect{\eta}$ can increase activation value and this growth is linearly related to weight vector $\vect{w}$. Satisfying the max norm constraint on adversarial perturbations, Goodfellow~\etal maximize the activation change by assigning $\vect{\eta} = \sign{(\vect{w})}$.

\IEEEpubidadjcol 
However, does $sign$ drive perturbations to the appropriate way? Probably not. Given typical attack method I-FGSM~\cite{ifgsm} as the baseline, in Figure~\ref{fig:hero} (a), we give a demonstration of the side-effect of $\sign$ in the $2D$ Euclidean space in the last step, where the origin $O'$ of the coordinate is close to $\ell_\infty$ $\epsilon$-ball. As shown in the figure, gradient $\vect{g}$ is extracted as two orthogonal direction, $\vect{u}$ and $\vect{v}$. Assuming original gradient is composed of $\vect{g} = 0.5\vect{u} + 0.1\vect{v}$. After $\sign$, the value is exceptional modified to $\frac{\epsilon}{T}\vect{u} + \frac{\epsilon}{T}\vect{v}$, having the included angle $\angle\theta=33.7^{\circ}$ between $\vect{g}$ and $\sign\left(\vect{g}\right)$. Obviously, after $sign$, the direction of the perturbation is not strictly following the original gradient. Although $\sign\left(\vect{g}\right)$ produces more perturbations under the $\ell_\infty$-norm scenario, it also introduces bias into the direction of perturbations. Since gradient is treated as the direction of fastest increase (untargeted) of $\mathcal{L}$ under the Euclidean feature space, the deviation in direction may make current gradient-based methods sub-optimal.  

Therefore in this paper, we review the state-of-the-art attacks and give the correction on deviation in direction. Firstly, we introduce Taylor expansion to give a systematic analysis. Secondly, based on the analysis, our approach, Fast Gradient Non-sign Method is proposed, which completely replaces $sign$ with a refined one. Thirdly, by modifying the original algorithms as little as possible, such seamless replacement has been shown to boost all methods in the exhaustive experiments significantly. Figure~\ref{fig:vis} provides the toy example of climbing the mountain with three strategies to show the advantages of our FGNM. As demonstrated in the figure, noticed that the optima are inside the $\epsilon$-ball, our two variants easily approach the optima. However, I-FGSM is stuck due to distorted perturbations. Specifically, when adopting I-FGSM, the trajectory is biased due to the limitation of $\sign$. It can not approach the optima and results in oscillation. Our I-\FGBMN holds the identical step size with I-FGSM in each iteration, but produces a more precise trajectory and is close to the optima. The I-\FGBMK considers the trade-off between magnitude and precision of perturbations in each step and gives the best result.

Our contribution is summarized up as three folds:

1) Comprehensive analysis is conducted on the $\sign$ operation of gradient-based methods, which shows the bias between original gradients and the actual perturbations. As a result, this gap may hinder the abilities of current methods to maximize $\mathcal{L}$ and therefore needs further correction.

2) A general and novel routine, Fast Gradient Non-sign Methods are proposed to correct the currently adopted sign routines. With comprehensive analysis, our methods can obtain less gap between gradients and actual perturbations than previous ones, which brings us ability to obtain performance boosts. 

3) Extensive experiments reveal the effectiveness of our methods. Specifically, for untargeted attacks, given normally trained models as black-box models, ours outperform state-of-the-art attacks by \textbf{27.5\%} at most and \textbf{11.3\%} on average. When transferring our adversarial examples to defense models, we still achieve a noticeable performance gain (\ie, \textbf{25.8\%} at most and \textbf{7.0\%} on average). For targeted black-box attacks against defense models, our FGNM surpasses vanilla methods by \textbf{8.3\%} on average and \textbf{18.5\%} at most.

\section{Related Works}
Adversarial attacks are a rising topic in recent advances of artificial intelligence. Although neural networks go deeper and deeper, researchers find they are still vulnerable to specific inputs. In general, adversarial attacks can be classified into three categories: white-box, gray-box, and black-box attacks, which indicates how much information attackers can exploit from the victim's models. White-box attacks have achieved noticeable attack performance since they are able to get all information from the victim's models~\cite{whitebosattack1,whitebosattack2}. For gray-box attacks, adversaries can only access the output logits or predictions, and adversarial examples are always crafted through huge amount queries~\cite{queryattack1,queryattack2}. When all information from the victim's models is unavailable, it becomes the most challenging black-box attack, and current methods always suffer from poor effectiveness. We mainly focus on black-box attacks.
\subsection{Black-box Attacks}
Because the decision boundaries of different models are similar, an important property of adversarial examples is their transferability, \ie, resultant adversarial examples generated from the white-box models are also effective for other models~\cite{transfer}. To tackle the poor performance of black-box attacks, several methods have been proposed to improve the transferability of adversarial examples~\cite{ifgsm,difgsm,sifgsm,mifgsm,tifgsm,pofgsm,PAAGAA,inter_transfer_TIP,UAP_AC_TIP}. For example, Goodfellow \etal \cite{fgsm} propose the fast gradient sign method to craft adversarial examples with just one step. I-FGSM~\cite{ifgsm} extends FGSM to an iterative attack and further improves the performance of white-box attacks at the cost of transferability. To further improve the transferability, by creating diverse patterns, DI-FGSM~\cite{difgsm} finds adversarial examples which transfer better to black-box models. MI-FGSM~\cite{mifgsm} employs momentum term into the perturbation generation and achieves noticeable performance gains. TI-FGSM~\cite{tifgsm} optimizes a perturbation over an ensemble of translated images and achieves translation-invariant attacks. SI-FGSM~\cite{sifgsm} crafts more transferable adversarial examples with the help of the scale copies of inputs.
\subsection{Defending against adversarial attacks}
Good transferability of adversarial examples enables black-box attacks feasible in physical world~\cite{physicalAttack1,physicalAttack2,physicalAttack4,physicalAttack5,physicalAttack6,physicalAttack7,physicalAttack8,physicalAttack9,physicalAttack10}. Consequently, a variety of defense methods have been proposed to evade the threat of adversarial attacks. Characterized by defense approach, there are several main defense methods. Gradient masking~\cite{mask1,mask2,mask3} which hinders attackers from obtaining true gradients, are mainly proposed to protect models from optimization-based attacks. Auxiliary detection models~\cite{defense_auxilary4} usually need to elaborate auxiliary models for predicting whether input is legitimate or adversarial. Other defense strategies will resort to preprocessing,~\eg, image transformations~\cite{defense_preprocess1,defense_preprocess2,defense_manifold_TIP}, denoising techniques~\cite{defense_prepocess3,defense_LRC_TIP}. Some statistical methods~\cite{defense_distribution} consider distribution of clean and adversarial samples as inconsistent. Therefore they usually detect adversarial examples with the help of statistics for distribution comparison. Defenses based on an ensemble of classifiers~\cite{ensemble} are approaches formed by several classifiers. Adversarial training, one of the most powerful defenses methods, trains models to be more robust to adversarial attacks by introducing adversarial examples into dataset~\cite{ensemble,adv_training}.

\section{Methodology}
In this paper, we mainly discuss non-targeted attacks, and their targeted version can be simply derived~\footnote{As adversarial examples generation pipelines of targeted and untargeted attacks are basically identical, later analysis of untargeted attacks is also adapted to targeted attacks. We omit to discuss targeted attacks for simplicity.}. Formally, given arbitrary input image $\vect{x} \in \mathbb{R}^{H\times W \times C}$ with ground-truth label $y$, non-targeted adversarial example $\vect{x}^{\mathit{adv}}$ aims at maximizing loss function ${\mathcal{L}\left(\vect{x}^\mathit{adv}, y\right)}$ under the ${\mathcal{\ell}_p}$ norm constraint:
\begin{equation}
    \label{eq:main}
    \begin{split}
        &\argmax_{\vect{x}^\mathit{adv}}{\mathcal{L}\left(\vect{x}^\mathit{adv}, y\right)}, \\
        &\mathit{s.t.} \norm{\vect{x}^\mathit{adv} - \vect{x}}_p \leq \epsilon.
    \end{split}
\end{equation}
To measure the human-imperceptibility of adversarial examples, following previous works ~\cite{mifgsm,tifgsm,difgsm,pofgsm, sifgsm,ifgsm}, $\ell_\infty$-norm is adopted in our work~\footnote{For case of $\ell_2$-norm please see~\ref{sec.discussion}}. Specifically, when $p=\infty$, {$\max\abs{\vect{x}^{\mathit{adv}}-\vect{x}}\leq{\epsilon}$}.

Some gradient-based attack methods~\cite{ifgsm,mifgsm,tifgsm} have been proposed to solve the optimization problem~(\ref{eq:main}), \eg, I-FGSM performs $T$-step attack with a small step size $\alpha = \epsilon / T$:
\begin{equation}
    \label{eq:I-FGSM}
    \begin{split}
        \vect{x}^{adv}_0 &= \vect{x},\;\vect{g}_t = \nabla_{\vect{x}^{adv}_t}\mathcal{L}\left(\vect{x}^{adv}_t, y\right),\\
        \vect{x}^\mathit{adv}_{t+1} &= \clip_\epsilon^{\vect{x}}\left(\vect{x}^{adv}_t + \alpha\cdot\sign\left(\vect{g}_t\right)\right), 0 \leq t < T,\\
    \end{split}
\end{equation}
where $\sign$ operation makes perturbations meet the $\ell_\infty$-norm bound as soon as possible, and clip operation is performed to make $\vect{x}^{adv}$ satisfy the $\epsilon$-ball of $\vect{x}$. Specifically:
\begin{equation}
    \label{eq:I-FGSM1}
    \begin{split}
        \sign\left(\cdot\right) &= \left\{
        \begin{array}{cc}
            +1 & \cdot > 0 \\
            0  & \cdot = 0 \\
            -1 & \cdot < 0
        \end{array}
        \right. \\
        \clip_\epsilon^{\vect{x}}\left(\cdot\right) = &\min\left(\max\left(\cdot, \vect{x} - \epsilon\right), \vect{x} + \epsilon\right).
    \end{split}
\end{equation}
Adopting basic pipeline of I-FGSM, other iterative methods are arranged into similar formulations, where the basic idea is generalized: performing method-specific manipulation on gradients $\vect{g}_t$ and then generating perturbations by $\sign$ operation.

\subsection{Taylor Expansion of Loss Function}
\label{talor}
To theoretically solve the optimization problem, \ie, Equation~(\ref{eq:main}), we derive the first-order Taylor polynomial of $\mathcal{L}$ at the point $\vect{x}_{T-1}^{adv}$ as:
\begin{equation}
    \label{eq:taylor-expand}
    \begin{split}
        \mathcal{L}\left(\vect{x}_{T}^\mathit{adv}, y\right)
        &=\mathcal{L}\left(\vect{x}_{T-1}^{adv}, y\right) \!+\! \left(\vect{x}_{T}^\mathit{adv}\!-\!\vect{x}_{T-1}^{adv}\right) \!\cdot\! \nabla{\mathcal{L}\left(\vect{x}_{T-1}^\mathit{adv},\! y\right)}\!+\!\mathcal{R}_{1}\\
        &=\mathcal{L}\left(\vect{x}, y\right)+\sum_{t=0}^{T-1}{(\vect{x}_{t+1}^\mathit{adv}-\vect{x}_t^{adv}) \cdot \vect{g}_t}+\mathcal{R}_{1},
    \end{split}
\end{equation}
where $\mathcal{R}_{1}$ is the remainder of first order and $\vect{g}_t = \nabla{\mathcal{L}\left(\vect{x}_{t}^\mathit{adv},\! y\right)}$. Under the $\ell_\infty$-norm constraint, we assume $\vect{x}_{t+1}^{adv} - \vect{x}_{t}^{adv}$ is so small that $\mathcal{R}_{1}$ can be neglected. By denoting ${\vect{x}_{t+1}^{adv}-\vect{x}_t^{adv}}=\vect{\delta}_t$, Equation~(\ref{eq:taylor-expand}) is rewritten as:
\begin{equation}
    \label{eq:taylor-expand1}
    \begin{split}
        \mathcal{L}\left(\vect{x}_{T}^\mathit{adv}, y\right)
        &=\mathcal{L}\left(\vect{x}, y\right)+\sum_{t=0}^{T-1}{\norm{\vect{\delta}_t}\norm{\vect{g}_t}\cos{\innerprod{\vect{\delta}_t, \vect{g}_t}}}. \\
    \end{split}
\end{equation}
Obviously, $\sum_{t=0}^{T-1}{\norm{\vect{\delta}_t}\norm{\vect{g}_t}\cos{\innerprod{\vect{\delta}_t, \vect{g}_t}}}$ is the key component for maximizing $\mathcal{L}\left(\vect{x}_{t+1}^\mathit{adv},y\right)$, where $\norm{\vect{\delta}_t}$, ${\vect{g}_t}$ is the magnitude of noises, gradient at iteration $t$. $\innerprod{\vect{\delta}_t, \vect{g}_t}$ is the included angle between $\vect{\delta}_t$ and $\vect{g}_t$. Although $\norm{\vect{g}_t}$ also affects the results, we treat it identical in different trajectories by assuming loss function is smooth and gradient does not change rapidly in a small region. 

\subsection{Drawbacks of Sign-based Attacks}
\begin{figure}[t]
    \begin{center}
       \includegraphics[width=0.72\columnwidth]{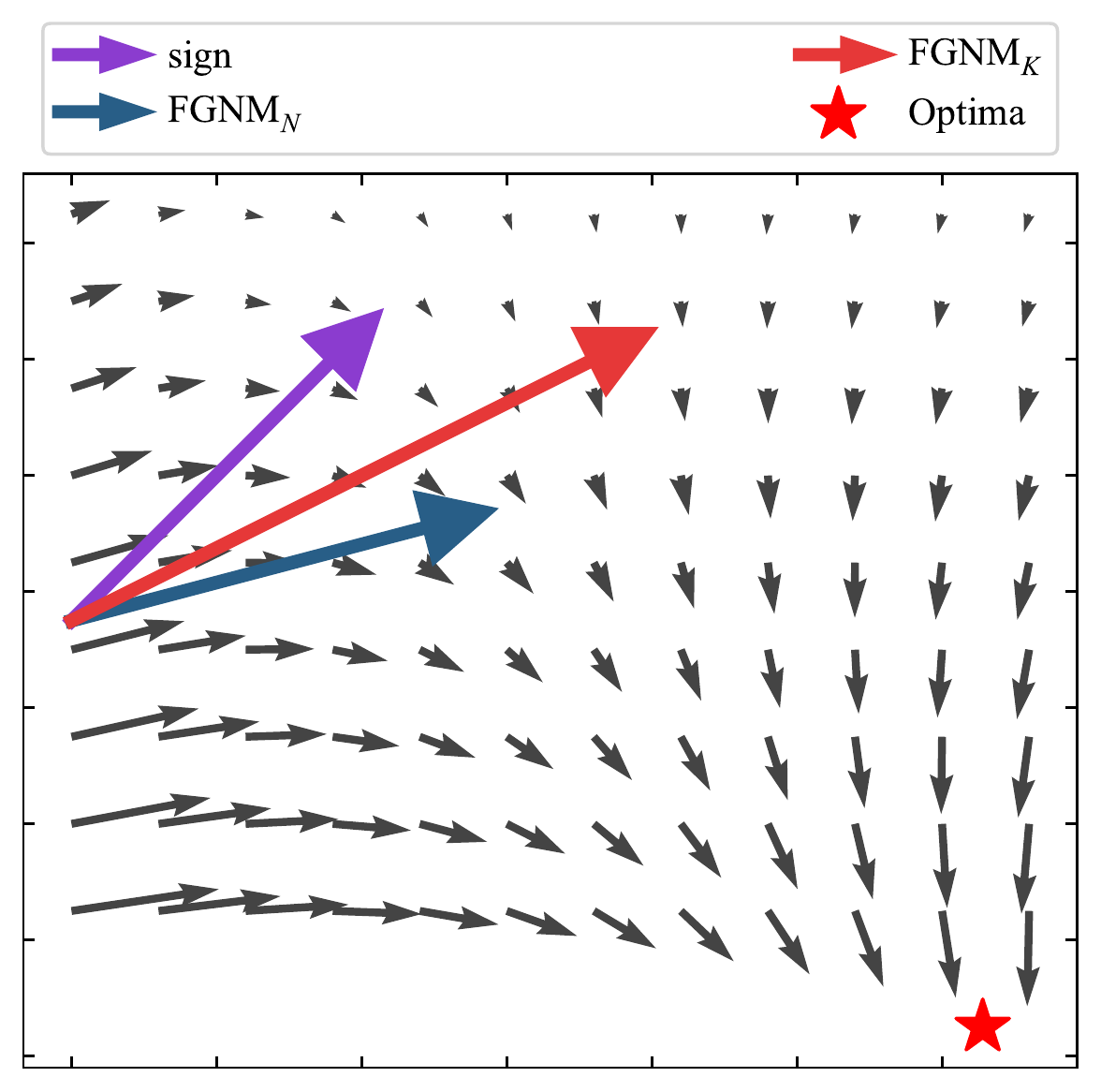}
    \end{center}
   \caption{Quiver plot for visualizing gaps between original gradients and actual perturbations in the demo scenario. To approach optima in the fastest way, a step's magnitude and direction are vital components. In this figure, our \FGBMN and \FGBMK take the right step at the starting point. However, the $\sign$ method goes far away instead.}
    \label{fig:local}
\end{figure}

As mentioned above, $\vect{g}_t$ indicates the direction of pulling $\vect{x}_t^{adv}$ to increase $\mathcal{L}$. However, considering that $\sign$ coarsely quantifies gradients to $\set{+1, -1, 0}$, there always exists a bias of direction between the perturbation and gradients. As Figure~\ref{fig:local} demonstrates, $\sign$ method pushes the point far away from the optima, which indicates the side-effect of this biased gradient manipulation. Based on the analysis in Sec.~\ref{talor}, in following sections we will explain why adopting $\sign$ will make Equation~(\ref{eq:taylor-expand1}) sub-optimal and give our approach.

For simplicity, we hypothesize $\vect{g_t}$ is non-zero. $\vect{x}_t^{adv}$ and $\vect{g}_t$ are interpreted as flattened vectors with dimension $D = H \times W \times C$:
\begin{equation}
    \begin{split}
        \vect{x}_t^{adv} &= \left[x^1, x^2, \cdots, x^D\right]_t, \\
        \vect{g}_t &= \left[\nabla_{x^1}, \nabla_{x^2}, \cdots, \nabla_{x^D}\right]_t.
    \end{split}
\end{equation}
To make the derivation simple, we select the typical sign-based attack: I-FGSM as our baseline. Then, $\vect{\delta}_t = \sign{\left(\vect{g}_t\right)}$ and $\cos{\innerprod{\vect{\delta}_t, \vect{g}_t}}$ is denoted as $\cos\theta_t$, where:
\begin{equation}
    \label{eq:cosineTheta}
    \begin{split}
        \cos{\theta}_t = \frac{\vect{g}_t \cdot \sign\left(\vect{g}_t\right)}{\norm{\vect{g}_t}\norm{\sign\left(\vect{g}_t\right)}}
        = \frac{\norm{\vect{g}_t}_1}{\sqrt{\norm{\vect{g}_t}_0}\norm{\vect{g}_t}}.
    \end{split}
\end{equation}
Here, $\norm{\vect{g}_t}_1 = \sum{\abs{\vect{g}_t^j}}$, $\norm{\vect{g}_t}_0 = D$. Due to the fact that $\norm{\vect{\cdot}}_1 \geq \norm{\vect{\cdot}}$ for any vector and $\norm{\vect{\cdot}}_1 = \norm{\vect{\cdot}}$ \emph{iff} the vector is all zero or it only has one non-zero entry, $\cos{\theta}$ ranges in:
\begin{equation}
    \frac{1}{\sqrt{D}} \textless{\cos{\theta}} \leq 1.
\end{equation}
Obviously,
\begin{equation}
    \label{eq:lambda}
    \begin{split}
        \norm{\vect{\delta}_t} = \alpha\sqrt{D}
    \end{split}
\end{equation}
\begin{equation}
    \alpha \textless{\norm{\vect{\delta}_t}\cos{\theta}} \leq \alpha\sqrt{D}.
\end{equation}
Two special cases will make $\cos{\theta} = 1$\eat{or $\vect{g}$ and $\sign\left(\vect{g}\right)$ are coincided}: a) $D = 1$, or b) All values of $\vect{g}_t$ are identical, which makes $\norm{\vect{g}_t}_1 = \sqrt{D}\norm{\vect{g}_t}$.

The derivation of $\theta_t$ indicates the gap between real gradients and actual perturbations at current iteration $i$. In fact, special cases mentioned above are nearly impossible to approach in practice, especially in high-dimensional space. Therefore the ideal maximum of $\norm{\vect{\delta}_t}\cos{\theta}_t$ is difficult to achieve, and we usually seek the sub-optimal solution of Equation~(\ref{eq:taylor-expand}).

\subsection{Fast Gradient Non-sign Methods}
\label{FGBM_methods}
\subsubsection{Fixed Scale Approach}
\label{sec.nmethods}
To reduce the side-effect caused by $\sign$, Iterative Fast Gradient Non-sign Method~(I-FGNM) and other variants are proposed. To distinguish our method, we denote the magnitude of our noise as $\norm{\vect{\delta'}}$ and the included angle as $\phi$. An intuitive solution is to perform a linear scale on the $\vect{g}_t$ so that $\norm{\vect{\delta'}} = \norm{\sign\left(\vect{g}_t\right)}$ and naturally $\cos\phi = 1 \geq \cos\theta$. To achieve this, a scale factor $\zeta$ is defined as:
\begin{equation}
    \label{eq:scaleN}
    \zeta = \frac{\norm{\sign(\vect{g}_t)}}{\norm{\vect{g}_t}}, \;\vect{\delta'_t} = \alpha\cdot\zeta\cdot\vect{g}_t.
\end{equation}


We call this simple solution as I-\FGBMN since the norm between $\vect{\delta'}$ and $\sign{\left(\vect{g}\right)}$ are equal.

Empirically in iterative algorithms, the step size $\alpha$ is much smaller than $\epsilon$. So in the first few steps, our noises are too small to be clipped. We have:
\begin{equation}
    \label{eq:noclippFGNM}
    \norm{\vect{\delta}_t}\cos{\theta}_t \leq \norm{\vect{\delta'_t}}\cos{\phi}_t = \alpha\sqrt{D}.
\end{equation}

The derivation shows that $\mathcal{L}$ calculated from our approach will always be greater or equal than the $\sign$ approach, according to Equation~(\ref{eq:noclippFGNM}) and (\ref{eq:taylor-expand1}) if there is no clipping. However, in the last few steps, the accumulated noises of previous steps make $\vect{x}^\mathit{adv}_t$ close to the $\epsilon$-ball, and it will be more likely to be clipped. In this scenario, the actual noises added in the $t$-th iteration will be:
\begin{equation}
    \label{eq:noiseOfB}
    \vect{x}_{t+1}^\mathit{adv} - \vect{x}_t^{adv} = \clip_\epsilon^{\vect{x}}\left(\vect{x}_t^{adv} + \alpha \cdot \vect{\delta'_t}\right) - \vect{x}_t^{adv}.
\end{equation}

Clipping introduces biases, \ie, $\norm{\vect{\delta'_t}}$ decreases and $\cos\phi_t$ will be smaller than $1$. It is complicated to judge the clipped $\vect{\delta'_t}$ and $\delta_t$ which one is better. And instead, we analyze it in a qualitative way. The clipped $\vect{\delta'_t}$ contains two parts: The non-clipped component and clipped component. If we only consider the non-clipped one, it still preserves the original gradients and performs linear scale. On the contrary, the clipped one only contains $-\epsilon$ or $+\epsilon$. So we treat it as \textbf{\textit{partially distorted}}. Different from $\sign{\left(\vect{g}_t\right)}$ which is \textbf{\textit{fully distorted}}, our mixed vector is more likely to be closed to $\vect{g}_t$ than $\sign\left(\vect{g}_t\right)$. Figure~\ref{fig:hero} demonstrates the analysis on clipping. As the figure shows, the $\vect{u}$-component of $\vect{\delta'}$ is reduced to $\epsilon$ while the $\vect{v}$-component of $\vect{\delta'}$ is preserved. The actual perturbation is denoted as $\tilde{\vect{\delta'}}$. Easily verified in the figure, the scaled and clipped result $\tilde{\vect{\delta'}}$ is closer to $\vect{g}_t$. Overall, $\sum_{t=0}^T{\norm{\tilde{\vect{\delta'_t}}}\cos{\phi_t}}$ is larger than $\sum_{t=0}^T{\norm{\vect{\delta}_t}\cos{\theta_t}}$.

\subsubsection{Adaptive Scale Approach}
\label{sec.kmethods}
Although I-\FGBMN is applicable for most cases by fixing $\norm{\vect{\delta'_t}}$, the attack performance is not always guaranteed due to clipped ratio. Empirically, a large clipped ratio hinders the performance of I-\FGBMN. To consider this ratio formally and comprehensively, we relax the constraint on $\norm{\vect{\delta'_t}}$ and compose a heuristic method I-\FGBMK, where the scale factor of I-\FGBMK is set by directly finding the proper value from all possible values:
\begin{equation}
\label{eq:scaleK}
    \begin{split}
        \mathit{scales} &= \operatorname{Sort}\left(\frac{1}{\abs{\vect{g}_t^k}}\right),\;\vect{g}_t^k \in \vect{g}_t, \\
        \zeta &= \mathit{scales}\left[K\right].
    \end{split}
\end{equation}
Specifically, $\mathit{scales}$ are calculated and sorted from $1 / \abs{\vect{g}_t^k}$, then $\zeta$ is picked by the $K$-th \textit{biggest} value of $scales$. As described before, two components (magnitude and direction of noise) essentially influence attack performance. For I-\FGBMK, there exists a trade-off between the two components, which is influenced by controlling $K$. Intuitively, larger $K$ leads to smaller magnitude, smaller clipped ratio, and larger $\cos\phi_t$. The Fast Gradient Non-sign Methods is summarized in Algorithm~\ref{alg.all}.




\IncMargin{1em}
\begin{algorithm}[!htbp]
    \caption{Algorithms of I-FGNM}
    \label{alg.all}
    \DontPrintSemicolon
    \SetAlgoLined
    \SetNoFillComment
    \SetKwInOut{Input}{\textbf{Input}}\SetKwInOut{Output}{\textbf{Output}}
	\Input{A classifier $\mathcal{F}$ with loss function $\mathcal{L}$; input image $\vect{x}$ with ground-truth label $y$; $T$ iterations; the size of perturbation $\epsilon$;
	}
	\Output{Adversarial example $\vect{x^{adv}}$ with $\norm{\vect{x}^{adv}-\vect{x}}_\infty\leq\epsilon$;
	}
	Initialize: ${\vect{x^{adv}_0}}=\vect{x}$, $\alpha = \epsilon/T$

	$t \leftarrow 0$;

    \While{$t < T$}
    {
        \tcc{generation loop of ${\vect{x^{adv}}}$}
        Input ${\vect{x^{adv}_t}}$ to $\mathcal{L}$ and obtain the gradient $\vect{g}_t=\nabla _{\vect{x}}\mathcal{L}\left(\vect{x}^{adv}_t,y\right)$;
        
        Calculate $\zeta$ by Equation~(\ref{eq:scaleN}) or Equation~(\ref{eq:scaleK});
        
        Update $\vect{x}^{adv}_{t+1}$ by our non-sign operation as
        $\vect{x}^{adv}_{t+1}=\vect{x}^{adv}_t+\alpha\cdot\zeta\cdot \vect{g}_t$;
        
        $t \leftarrow t + 1$;
    }
\end{algorithm}
\DecMargin{1em}

\subsection{Discussion}
\label{sec.discussion}
Our Fast Gradient Non-sign Methods are delivered with two variants, \FGBMN and \FGBMK. The former generates noises $\vect{\delta'}$ by linearly scaling from $\vect{g}$ and keeping the magnitude same as $\norm{\sign{\left(\vect{g}_t\right)}}$ with a scale $\zeta$. The latter relaxes the magnitude to consider the trade-off between $\norm{\vect{\delta'}}$ and $\cos\phi$. Moreover, the replacement is seamless and universal. Any current FGSM-based attacks are applicable to become FGNM-based attacks, \eg, MI-FGSM, TI-FGSM, DI-FGSM, SI-FGSM, and the computational overhead is negligible.

Noticed that there also exists another style of attack whose constraint is defined as $\norm{\vect{x^{adv}}-\vect{x}} \leq \epsilon$ ($\ell_2$ attacks). We need to clarify the difference between our FGNM and $\ell_2$ attacks. Firstly, since the constraint is changed, $\ell_2$ attacks are not troubled by the $\ell_\infty$ constraint, which causes the gradients to be distorted. Secondly, our approaches essentially consider reducing the side-effect of $\sign$ operation to leverage the performance of $\ell_\infty$ attacks finally. Therefore, we believe our approaches have advantages among the $\ell_\infty$ attacks.


\begin{table}[t]
\caption{Hyper-parameters adopted in our experiments for MI-FGSM, DI-FGSM, TI-BIM, and SI-FGSM, respectively. For a fair comparison, our methods use identical hyper-parameters.}
\label{tab:hypers}
\centering
\resizebox{1\columnwidth}{!}{
\begin{tabular}{c|c|cll}\hline\hline
Methods             & Hyper-parameters    & \multicolumn{3}{c}{Value}                               \\ \hline
MI-FGSM                  & Momentum $\mu$                 & \multicolumn{3}{c}{1.0}                                   \\ \hline
DI-FGSM                 & Transform probability $p$                  & \multicolumn{3}{c}{0.7}                                   \\ \hline
\multirow{2}{*}{TI-BIM} & \multirow{2}{*}{Kernel length $W$} & \multicolumn{1}{c|}{Untargeted} & \multicolumn{2}{l}{15} \\ \cline{3-5} 
                    &                    & \multicolumn{1}{c|}{Targeted}   & \multicolumn{2}{l}{5} \\ \hline
SI-FGSM                  & Number of the scale copies $c$                  & \multicolumn{3}{c}{5}                                  \\ \hline\hline
\end{tabular}%
}
\end{table}

\section{Experiments}
\label{Sec.Experiments}
To confirm our analysis and compare with currently state-of-the-art attacks, extensive experiments are conducted in non-targeted and targeted $\ell_\infty$ attack scenario.

\subsection{Setup}

\paragraph{Dataset} The ImageNet-compatible dataset in the NIPS 2017 adversarial competition\footnote{https://www.kaggle.com/c/nips-2017-non-targeted-adversarial-attack} is chosen to evaluate our methods and state-of-the-art attacks. This dataset contains $1,000$ images and is widely used in previous works~\cite{mifgsm,tifgsm,pifgsm,sifgsm}.

\paragraph{Models} To give comprehensive statistical analysis, experiments are conducted on eight models, including five normally trained models~(NT): Inception V3~(\inca)~\cite{incv3}, Inception V4~(\incb)~\cite{incv4}, ResNet152 V2~(\res)~\cite{res152}, Inception-Resnet V2~(\inre)~\cite{inc-res} and DenseNet 161~(\den)~\cite{den161}, and three ensemble adversarially trained models~(EAT): \incaE, \incaE and \inreE~\cite{ensemble}. 

\paragraph{Baselines} As described above, our FGNM are applicable for any FGSM-based methods by directly replacing $\sign$ with scale $\zeta$. In this paper, six currently popular FGSM-based attacks are selected as baselines: I-FGSM~\cite{ifgsm}, MI-FGSM~\cite{mifgsm}, DI-FGSM~\cite{difgsm}, TI-BIM~\cite{tifgsm}, PoI-FGSM~\cite{pofgsm} and SI-FGSM~\cite{sifgsm}. Our boosted versions are denoted as I-FGNM, MI-FGNM, DI-FGNM, TI-BNIM, PoI-FGNM, SI-FGNM, respectively. Both two variants (\FGBMN and \FGBMK) are evaluated.

\paragraph{Hyper-parameters} To make a fair comparison, we follow the previous works~\cite{mifgsm,difgsm,tifgsm,sifgsm} to set hyper-parameters. For the generalized parameters, we set $\epsilon = 16$ with pixel-level ranges in $0\smallsim255$. And for untargeted iterative method, the total iteration number $T$ and step size $\alpha$ are 10 and $\epsilon/10 = 1.6$, respectively. For targeted iterative method, they are 20 and $\epsilon/20 = 0.8$. As for method-specific hyper-parameters, we adopt the original value reported in their paper, as shown in Table \ref{tab:hypers}. Note that our methods will not modify these hyper-parameters.


\begin{figure*}[!htbp]
    \begin{center}
       \includegraphics[width=\textwidth]{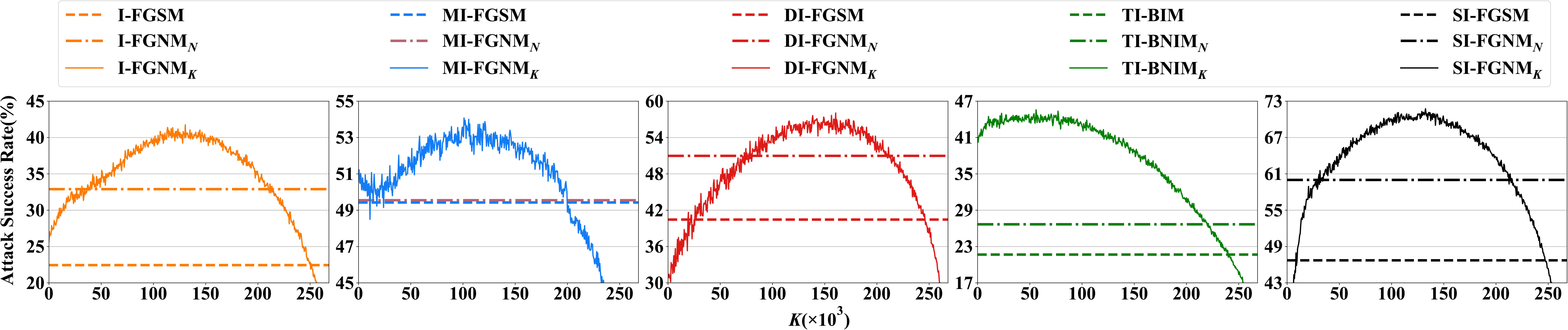}
    \end{center}
   \caption{Average untargeted attack success rate comparisons between $\sign$ methods and our {\FGBMK} variants on normally trained models. We employ \inca as white-box and compute average attack success rate of other four black-box models (\incb, \res, \inre, and \den) \wrt $K$. Results of sign-based methods and {\FGBMN} variants are also reported as dashed lines and dash-dotted lines, respectively.}
    \label{fig:Kselect_NT}
\end{figure*}

\begin{figure*}[t]
    \begin{center}
       \includegraphics[width=\textwidth]{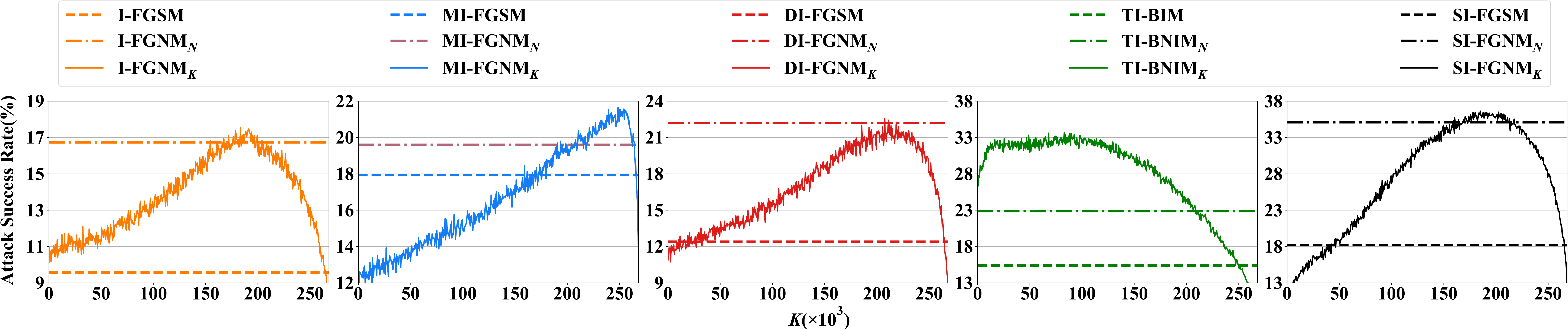}
    \end{center}
   \caption{Average untargeted attack success rate comparisons between $\sign$ methods and our {\FGBMK} variants on adversarially trained models. We employ \inca as white-box and compute average attack success rate of three black-box models (\incaE, \incbE, \inreE) \wrt $K$. Unlike NT, the best results present in the later area in most cases. Results of sign-based methods and {\FGBMN} variants are also reported as dashed lines and dash-dotted lines, respectively.}
    \label{fig:Kselect_EAT}
\end{figure*}

\begin{figure}[t]
    \begin{center}
       \includegraphics[width=1\columnwidth]{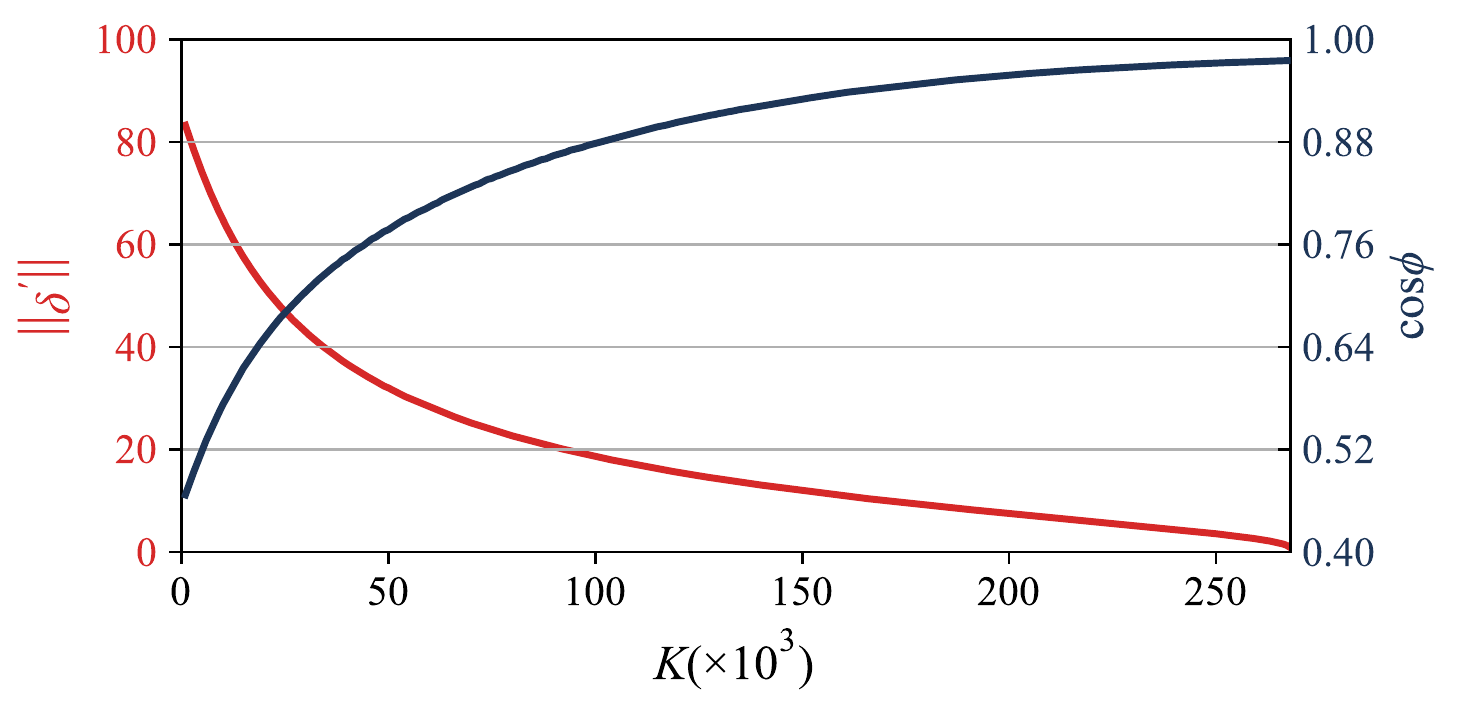}
    \end{center}
   \caption{Evaluation on average $\cos\phi$ and $\norm{\delta'}$ \wrt $K$ to show the changes of magnitude and direction of added perturbations. Obviously, $\norm{\delta'}$ decreases when $K$ increases, and $\cos\phi$ goes up in contrast. The trends confirm our analysis of magnitude and direction of noises.}
    \label{fig:cos_lambda}
\end{figure}

\begin{table*}[t]
\caption{Quantitative comparisons between state-of-the-art attacks on normally trained models. Specifically, Inc-v3, Inc-v4, Res152 and IncRes are adopted to craft untargeted adversarial examples respectively. \enquote{*} denotes the white-box results while others are black-box results. 
}
\label{tab:attack_s_NT}
\centering
\resizebox{0.81\textwidth}{!}{
\begin{tabular}{c|c|ccccc|c|ccccc}\hline
                  Models & Attacks  & \inca  & \incb  & \res   & \inre   & \den     &                                                   Models   & \inca  & \incb  & \res   & \inre  & \den \\ \hline\hline
\multirow{15}{*}{\inca}  & I-FGSM    &\textbf{100.0*}   &29.4    &18.9    &20.3     &21.2               & \multirow{15}{*}{\incb}       &42.0    &\textbf{100.0*}   &22.6   &25.8    &24.4        \\
                         & I-\FGBMN (ours)  &\textbf{100.0*}   &41.9    &28.2    &33.4 &28.1      &                                   &51.5    &\textbf{100.0*}    &32.2    &35.5    &31.0       \\
                         & I-\FGBMK (ours)  &\textbf{100.0*}   &\textbf{48.7}    &\textbf{32.7}  &\textbf{44.2} & \textbf{36.5}      &         &\textbf{63.0}    &\textbf{100.0*}   &\textbf{40.4}    &\textbf{46.0}    &\textbf{44.3}\\ \cline{2-7} \cline{9-13}
                         & MI-FGSM   &\textbf{100.0*}   &55.1    &43.0    &50.9     &48.7      &                                        &70.8    &\textbf{100.0*}   &52.2    &59.0    &55.5        \\
                         & MI-\FGBMN (ours) &\textbf{100.0*}   &56.2    &42.8    &52.3     &46.9      &                                        &69.7    &\textbf{100.0*}   &50.0    &58.2    &52.3        \\
                         & MI-\FGBMK (ours) &\textbf{100.0*} &\textbf{59.0} &\textbf{45.9} &\textbf{57.9} &\textbf{51.1} &                     &\textbf{74.3}    &\textbf{100.0*}   &\textbf{54.0}    &\textbf{63.3}    &\textbf{56.5}\\ \cline{2-7} \cline{9-13}
                         & DI-FGSM   &99.9*    &54.9    &33.6    &41.7     &31.6      &                                                 &64.9    &\textbf{100.0*}   &39.1    &49.4    &37.2        \\
                         & DI-\FGBMN (ours) &99.7*    &65.3    &43.7    &55.1     &39.8      &                                        &73.7    &\textbf{100.0*}    &47.8   &59.4    &45.2        \\
                         & DI-\FGBMK (ours) &\textbf{100.0*}   &\textbf{67.3} &\textbf{45.6} &\textbf{62.0} &\textbf{47.0}  &                           &\textbf{78.4}    &\textbf{100.0*}   &\textbf{55.0}    &\textbf{66.4}    &\textbf{54.5}\\ \cline{2-7} \cline{9-13}
                         & TI-BIM    &99.9*    &28.1    &16.7    &13.5     &28.4      &                                                 &35.2    &\textbf{100.0*}   &20.4    &17.0    &31.8        \\
                         & \TIBBIMN (ours) &\textbf{100.0*}    &25.9    &23.3    &23.9    &33.6      &                                         &46.3    &\textbf{100.0*}    &27.4    &26.7    &40.2        \\
                         & \TIBBIMK (ours) &\textbf{100.0*} &\textbf{47.4} &\textbf{33.6} &\textbf{35.9} &\textbf{47.6}  &                     &\textbf{58.2}     &\textbf{100.0*}    &\textbf{35.6}    &\textbf{39.4}    &\textbf{54.7}        \\ \cline{2-7} \cline{9-13}
                         & SI-FGSM   &\textbf{99.0*}   &53.8    &40.4    &47.8     &44.9      &                                                 &72.6    &\textbf{100.0*}   &51.0    &57.0    &57.2        \\
                         & SI-\FGBMN (ours) &\textbf{99.0*}   &{67.7}    &52.7    &{63.2}    &56.4     &                                                   &{78.0}    &\textbf{100.0*}   &{61.3}    &{70.8}    &{65.1}       \\
                         & SI-\FGBMK (ours) &\textbf{99.0*}     &\textbf{76.6}    &\textbf{64.1}    &\textbf{73.2}     &\textbf{67.0}      &  &\textbf{86.4}    &\textbf{100.0*}   &\textbf{71.9}    &\textbf{77.9}    &\textbf{76.6}        \\ \hline\hline
\multirow{15}{*}{\res}   & I-FGSM    &31.4    &25.1    &\textbf{99.5*}    &17.0     &23.8              & \multirow{15}{*}{\inre}        &46.2   &38.6    &27.6    &\textbf{100.0*}    &28.1        \\
                         & I-\FGBMN (ours)  &40.8    &35.3    &99.1*    &30.5     &34.3        &                                               &56.6   &49.3    &35.7    &99.9*    &35.8        \\
                         & I-\FGBMK (ours)  &\textbf{52.0} &\textbf{46.6} &\textbf{99.5*} &\textbf{42.4} &\textbf{44.9}        &               &\textbf{65.6}   &\textbf{53.8}    &\textbf{39.0}    &\textbf{100.0*}   &\textbf{43.5} \\ \cline{2-7} \cline{9-13}
                         & MI-FGSM   &56.1    &50.8    &\textbf{99.5*}    &47.2     &54.6        &                                               &76.4   &67.8    &{57.0}    &\textbf{100.0*}   &60.3        \\
                         & MI-\FGBMN (ours) &55.7    &49.6    &\textbf{99.5*}    &46.2     &52.7        &                                               &75.2   &65.6    &55.2    &\textbf{100.0*}   &58.3        \\
                         & MI-\FGBMK (ours) &\textbf{62.6}    &\textbf{56.7} &\textbf{99.5*} &\textbf{53.9}&\textbf{58.0}      &               &\textbf{77.0}   &\textbf{69.2}    &\textbf{58.6}    &\textbf{100.0*}   &\textbf{61.2}  \\ \cline{2-7} \cline{9-13}
                         & DI-FGSM   &62.1    &55.6    &99.1*    &49.4     &47.4        &                                               &70.6   &66.7    &48.1    &99.5*    &45.1        \\
                         & DI-\FGBMN (ours) &69.9    &64.2    &99.3*    &58.7     &56.5        &                                               &76.9   &74.3    &56.3    &99.6*    &51.9        \\
                         & DI-\FGBMK (ours) &\textbf{76.2}    &\textbf{70.1}    &\textbf{99.6}*    &\textbf{67.1}     &\textbf{64.8}        &  &\textbf{80.6}   &\textbf{76.8}    &\textbf{56.8}    &\textbf{98.8*}    &\textbf{57.8} \\ \cline{2-7} \cline{9-13}
                         & TI-BIM    &24.7    &21.3    &99.2*    &11.5     &30.9        &                                               &44.0   &40.2    &26.3    &99.4*    &41.0        \\
                         & \TIBBIMN (ours) &33.1    &27.8    &99.0*    &19.1     &35.9        &                                                &51.0   &45.4    &30.5    &\textbf{99.6*}    &46.5        \\
                         & \TIBBIMK (ours) &\textbf{47.0}    &\textbf{41.6}&\textbf{99.5*}&\textbf{33.3}&\textbf{54.2} &                       &\textbf{63.0}   &\textbf{56.5}    &\textbf{43.1}    &99.0*   &\textbf{61.6} \\ \cline{2-7} \cline{9-13}
                         & SI-FGSM   &43.6    &40.0    &\textbf{99.8*}    &30.6     &44.6        &                                      &72.8   &63.9    &52.7    &99.9*    &58.8        \\
                         & SI-\FGBMN (ours) &53.9    &48.9    &99.7*    &53.9     &56.3        &                                               &81.3   &73.5    &64.6    &99.9*    &67.7        \\
                         & SI-\FGBMK (ours) &\textbf{66.0}    &\textbf{62.6}   &\textbf{99.8*}    &\textbf{58.1}     &\textbf{68.6}        &           &\textbf{86.6}   &\textbf{80.6}    &\textbf{70.7}    &\textbf{100.0*}   &\textbf{75.4}       \\\hline
\end{tabular}
}
\end{table*}

\begin{table*}[t]
\centering
\caption{Quantitative comparisons between state-of-the-art attacks on adversarially trained models (EAT). \inca, \incb, \res, and \inre are adopted as white-box models to generate untargeted adversarial examples, while EAT are used to test transferability of adversarial examples.} 
\label{tab:attack_s_EAT}
\resizebox{0.96\textwidth}{!}{
\begin{tabular}{c|ccc|ccc|ccc|ccc}\hline
           & \multicolumn{3}{c|}{\inca} & \multicolumn{3}{c|}{\incb} & \multicolumn{3}{c|}{\res} & \multicolumn{3}{c}{\inre} \\
Attacks    &\incaE  &\incbE   &\inreE  &\incaE       &\incbE       &\inreE     &\incaE       &\incbE       &\inreE      &\incaE       &\incbE       &\inreE        \\\hline\hline
I-FGSM    &12.1     &12.6    &4.6        &12.0         &12.0         &5.8        &12.3         &13.4         &6.7         &13.3         &12.8         &8.9        \\
I-\FGBMN (ours)  &\textbf{18.9}     &\textbf{20.8}    &10.5        &18.7         &{19.5}         &{11.6}       &21.7         &22.5         &{15.5}        &{26.2}         &{23.8}         &{18.4}        \\
I-\FGBMK (ours)  &{18.5}&{20.1}&\textbf{10.6}        &\textbf{19.9}         &\textbf{20.7}         &\textbf{12.2}       &\textbf{22.4}         &\textbf{23.9}         &\textbf{16.1}       &\textbf{26.3}         &\textbf{24.2}         &\textbf{19.2}        \\\hline
MI-FGSM   &22.2            &21.2     &10.4    &24.6         &22.4         &13.4       &24.7         &26.7         &15.7        &31.2         &28.6         &20.7        \\
MI-\FGBMN (ours) &\textbf{22.8}   &23.2     &\textbf{12.8}    &\textbf{26.3}         &\textbf{26.4}         &\textbf{16.1}       &{31.0}         &\textbf{30.0}         &\textbf{19.0}        &\textbf{36.5}   &\textbf{32.1}&\textbf{25.7}        \\
MI-\FGBMK (ours) &21.9    &\textbf{23.3}   &{11.9}    &25.3         &25.2         &14.7       &\textbf{31.2}         &28.7         &18.4        &34.6         &31.5         &23.0        \\\hline
DI-FGSM   &14.2     &15.7     &7.3     &14.6         &16.8         &9.4        &21.0         &20.9         &12.4        &20.8         &18.7         &12.3        \\
DI-\FGBMN (ours) &\textbf{24.7}     &\textbf{27.0}     &\textbf{14.9}    &\textbf{27.3}         &{26.0}         &{17.1}       &\textbf{38.1}         &\textbf{36.2}         &{26.4}        &\textbf{38.3}         &{34.7}         &\textbf{28.5}       \\
DI-\FGBMK (ours) &{24.3}&{26.7}&{13.4}    &27.0         &\textbf{26.7}        &\textbf{17.3}       &37.7         &35.1         &\textbf{26.6}        &37.8         &\textbf{35.0}         &27.1        \\\hline
TI-BIM    &17.5     &18.0     &10.7    &17.2         &20.4         &11.8       &18.5         &18.7         &12.6        &25.4         &25.8         &21.1        \\
\TIBBIMN (ours) &25.6     &25.9     &17.1    &28.5         &28.3         &19.8       &25.3         &25.4         &19.7        &37.4         &35.9         &30.6        \\
\TIBBIMK (ours) &\textbf{27.3}&\textbf{27.3}&\textbf{18.8}    &\textbf{29.1}         &\textbf{28.4}         &\textbf{20.4}       &\textbf{27.9}         &\textbf{29.0}         &\textbf{21.3}        &\textbf{37.7}         &\textbf{37.0}         &\textbf{33.8}        \\\hline
SI-FGSM   &21.7     &22.5     &10.4    &27.3         &28.5        &16.9       &20.2         &18.7         &12.3        &30.0         &28.2         &21.6       \\
SI-\FGBMN (ours) &38.8     &41.1     &\textbf{25.4}    &{49.6}         &{48.5}         &\textbf{33.2}       &37.0         &34.6         &27.3        &54.3         &48.3         &45.7        \\
SI-\FGBMK (ours) &\textbf{40.0}     &\textbf{42.4}     &{25.3}    &\textbf{50.9}         &\textbf{49.0}        &{32.2}        &\textbf{37.9}         &\textbf{37.8}         &\textbf{27.4}        &\textbf{55.8}         &\textbf{50.5}         &\textbf{46.4}        \\\hline
\end{tabular}
}
\end{table*}

\subsection{Effectiveness of $K$}
\label{sec.Kstudy}
In this section, we firstly study the effectiveness of $K$ since it represents the trade-off between magnitude and bias of noises. 

To find an appropriate $K$, given Inc-v3 as white-box models and $K$ ranging from $1$ to $299\times299\times3$ with interval $500$, we firstly craft untargeted adversarial examples using all \FGBMK-based methods and then report the average attack success rate against black-box NT and EAT in Figure~\ref{fig:Kselect_NT} and Figure~\ref{fig:Kselect_EAT} respectively. We also plot the performance of sign-based methods and our \FGBMN methods, which will be discussed in the next sections.
As demonstrated from the vast majority of \eat{convex} concave curves in Figure~\ref{fig:Kselect_NT} and Figure~\ref{fig:Kselect_EAT}. As for NT, we give a unified optimal $K$ for all methods $K=120,000$. For EAT, $K=200,000$.

A natural question is that why most of the curves corresponding to our \FGBMK in Figure~\ref{fig:Kselect_NT} and Figure~\ref{fig:Kselect_EAT} are concave. 
In particular, we pick I-\FGBMK to answer this question. As described before, magnitude and bias of noises play a key role in performing successful attack. Therefore, a potential reason for the concave trends is analysis of these two components in Sec~\ref{sec.nmethods}. To confirm the analysis 
in Sec~\ref{sec.nmethods}, we measure these two values on I-\FGBMK. To obtain actual noises added in the $i$-th iteration, we compute $\vect{\delta}'_t = \vect{x}^\mathit{adv}_{t+1} - \vect{x}^\mathit{adv}_{t}$. Then, magnitude is computed by $\norm{\vect{\delta}'_t}$ and bias of direction is computed by $\cos\phi_t = \frac{\innerprod{\vect{\delta}'_t, \vect{g}_t}}{\norm{\vect{\delta}'_t}\norm{\vect{g}_t}}$ and they are averaged among whole iterations. The results with respect to $K$ are placed in Figure~\ref{fig:cos_lambda}. This figure shows our idea for presenting the trends of $\norm{\vect{\delta'}}$ and $\cos\phi$ when varying $K$. Particularly, when $K$ increases, the scale $\zeta$ decreases, and therefore the generated perturbations will be smaller. Meanwhile, smaller perturbations leads to a smaller clipped ratio, and instead preserves the direction. So the actual $\norm{\vect{\delta'}}$ goes down while $\cos\phi$ goes up. Naturally, there exists a maximum value of $\norm{\vect{\delta'}}\cos\phi$.
\subsection{Comparisons with State-of-the-Art Attacks}
\subsubsection{Performance of untargeted attacks}
After finding the optimal $K$, comprehensive experiments are conducted further to evaluate the effectiveness of our methods against NT and EAT under untargeted setting. 
\paragraph{Attack against Normally Trained Models}
In this section, we study the vulnerability of NT w.r.t different algorithms, including \FGBMN-based, \FGBMK-based, and their corresponding FGSM-based methods. Adversarial examples are crafted via given white-box models by different methods. 

\begin{table*}[t]
\caption{Quantitative comparisons between different combinations of methods on normally trained models. Untargeted aversarial examples are crafted via Inc-v3, Inc-v4, Res152 and IncRes. The rest of models are used to verify transferability. \enquote{*} denotes the white-box results while others are black-box results.
}
\label{tab:attack_c_NT}
\centering
\resizebox{0.85\textwidth}{!}{
\begin{tabular}{c|c|ccccc|c|ccccc}\hline
                  Models & Attacks  & \inca  & \incb  & \res   & \inre   & \den     &                                                   Models   & \inca  & \incb  & \res   & \inre  & \den \\ \hline\hline
\multirow{15}{*}{\inca}  & DTI-FGSM    &97.3*   &45.1    &25.5    &25.3     &41.0               & \multirow{15}{*}{\incb}       &35.0    &\textbf{100.0*}   &20.7   &17.1    &31.7        \\
                         & DTI-\FGBMN (ours)  &97.5*   &52.0    &31.9    &36.9 &46.7      &                                   &44.8    &\textbf{100.0*}    &25.8    &26.6    &40.0       \\
                         & DTI-\FGBMK (ours)  &\textbf{98.0*}   &\textbf{62.5}    &\textbf{42.5}  &\textbf{47.4} & \textbf{61.0}      &         &\textbf{58.6}    &\textbf{100.0*}   &\textbf{36.5}    &\textbf{41.7}    &\textbf{54.4}\\ \cline{2-7} \cline{9-13}
                         & DSI-FGSM   &\textbf{100.0*}   &83.9    &67.6    &76.0     &70.6      &                                        &88.2    &99.6*   &73.3    &81.6    &75.9        \\
                         & DSI-\FGBMN (ours) &\textbf{100.0*}   &87.5    &75.1    &83.2     &77.1      &                                        &90.9    &\textbf{100.0*}   &77.1    &87.2    &81.3        \\
                         & DSI-\FGBMK (ours) &\textbf{100.0*} &\textbf{90.9} &\textbf{80.4} &\textbf{87.5} &\textbf{83.5} &                     &\textbf{92.9}    &\textbf{100.0*}   &\textbf{83.7}    &\textbf{89.3}    &\textbf{87.9}\\ \cline{2-7} \cline{9-13}
                         & DTSI-FGSM   &99.7*    &68.6    &46.3    &53.5     &74.0     &                                                 &74.3    &99.9*   &48.7    &60.1    &77.5        \\
                         & DTSI-\FGBMN (ours) &99.9*    &74.5    &52.6    &62.2     &79.9      &                                                &77.6    &\textbf{100.0*}    &53.9   &66.7    &80.6        \\
                         & DTSI-\FGBMK (ours) &\textbf{100.0*}   &\textbf{82.1} &\textbf{61.5} &\textbf{69.4} &\textbf{86.3}  &                 &\textbf{84.1}    &\textbf{100.0*}   &\textbf{63.0}    &\textbf{71.1}    &\textbf{87.8}\\ \cline{2-7} \cline{9-13}
                         & MDTSI-FGSM    &99.5*    &77.2    &59.9    &63.4     &84.0      &                                              &80.3    &99.7*   &60.5    &67.4    &84.7        \\
                         & MDTSI-\FGBMN (ours) &99.7*    &75.3    &58.8    &62.4    &83.8      &                                       &78.3    &99.9*    &60.0    &66.1    &84.6        \\
                         & MDTSI-\FGBMK (ours) &\textbf{99.9*} &\textbf{80.4} &\textbf{64.6} &\textbf{67.7} &\textbf{88.1}  &                  &\textbf{83.1}     &\textbf{100.0*}    &\textbf{65.0}    &\textbf{71.5}    &\textbf{88.1}       \\ \hline\hline
\multirow{15}{*}{\res}   & DTI-FGSM    &45.1    &40.4    &97.9*    &30.6     &53.2              & \multirow{15}{*}{\inre}        &42.3   &39.3    &25.8    &99.4*    &40.0        \\
                         & DTI-\FGBMN (ours)  &52.2    &51.3   &98.4*    &42.4     &59.2        &                                               &51.2   &45.3    &31.0    &99.5*    &46.7       \\
                         & DTI-\FGBMK (ours)  &\textbf{63.9} &\textbf{59.8} &\textbf{99.2*} &\textbf{53.5} &\textbf{73.5}        &               &\textbf{63.4}   &\textbf{58.2}    &\textbf{42.6}    &\textbf{99.7*}   &\textbf{50.2} \\ \cline{2-7} \cline{9-13}
                         & DSI-FGSM   &79.9    &75.0    &\textbf{99.9*}    &72.7     &79.5        &                                               &91.7   &88.0    &81.9    &99.9*   &82.7        \\
                         & DSI-\FGBMN (ours) &83.9    &81.2   &\textbf{99.9*}    &79.8     &83.7        &                                               &92.6   &90.4    &83.7    &99.8*   &84.8        \\
                         & DSI-\FGBMK (ours) &\textbf{88.8}    &\textbf{86.9} &\textbf{99.9*} &\textbf{84.8}&\textbf{89.6}      &               &\textbf{94.6}   &\textbf{92.9}    &\textbf{87.3}    &\textbf{100.0*}   &\textbf{88.8}  \\ \cline{2-7} \cline{9-13}
                         & DTSI-FGSM   &61.3    &58.9    &98.8*    &54.1     &79.6        &                                               &81.4   &80.6    &65.6    &98.9*    &87.4        \\
                         & DTSI-\FGBMN (ours) &66.8    &66.7    &99.0*    &61.6     &81.4        &                                               &83.6   &81.0    &68.2    &99.3*    &88.4        \\
                         & DTSI-\FGBMK (ours) &\textbf{76.7}    &\textbf{74.0}    &\textbf{99.9}*    &\textbf{69.4}     &\textbf{89.5}        &  &\textbf{88.1}   &\textbf{87.3}    &\textbf{76.0}    &\textbf{99.8*}    &\textbf{92.9} \\ \cline{2-7} \cline{9-13}
                         & MDTS-FGSM    &67.3    &65.3    &98.2*    &60.7     &83.6        &                                               &81.2   &79.5    &70.3    &98.1*    &88.8        \\
                         & MDTSI-\FGBMN (ours) &67.3    &64.1    &98.9*    &61.3     &83.1        &                                                &82.1   &79.8    &69.1    &98.7*    &88.1        \\
                         & MDTSI-\FGBMK (ours) &\textbf{74.9}    &\textbf{69.1}&\textbf{99.4*}&\textbf{65.7}&\textbf{88.4} &                       &\textbf{85.6}   &\textbf{83.9}    &\textbf{74.7}    &\textbf{99.4*}   &\textbf{93.0}    \\\hline\hline
\end{tabular}
}
\end{table*}

\begin{table*}[t]
\centering
\caption{Quantitative comparisons between different combinations of methods on adversarially trained models (EAT). \inca, \incb, \res, and \inre are adopted as white-box models to generate untargeted adversarial examples, while EAT are used to test transferability of adversarial examples.} 
\label{tab:attack_c_EAT}
\resizebox{1\textwidth}{!}{
\begin{tabular}{c|ccc|ccc|ccc|ccc}\hline
           & \multicolumn{3}{c|}{\inca} & \multicolumn{3}{c|}{\incb} & \multicolumn{3}{c|}{\res} & \multicolumn{3}{c}{\inre} \\
Attacks    &\incaE  &\incbE   &\inreE  &\incaE       &\incbE       &\inreE     &\incaE       &\incbE       &\inreE      &\incaE       &\incbE       &\inreE        \\\hline\hline
DTI-FGSM    &23.2     &25.6    &14.2        &17.1         &20.4         &11.1        &33.8         &33.9         &24.1         &25.2         &25.4         &20.8        \\
DTI-\FGBMN (ours)  &34.3     &36.2    &22.7        &26.4         &{28.8}         &{18.4}       &45.5         &44.4         &{36.2}        &{36.3}         &{35.5}         &{32.4}        \\
DTI-\FGBMK (ours)  &\textbf{35.9} &\textbf{36.8}&\textbf{23.7}        &\textbf{28.1}         &\textbf{29.7}         &\textbf{21.2}       &\textbf{49.5}         &\textbf{48.4}         &\textbf{41.5}       &\textbf{39.7}         &\textbf{38.1}         &\textbf{25.5}        \\\hline
DSI-FGSM   &31.9            &31.0     &18.2    &42.1         &40.9         &26.2       &39.1         &36.5         &25.5        &49.4         &45.5         &37.4        \\
DSI-\FGBMN (ours) &\textbf{55.5}   &\textbf{55.9}     &\textbf{36.9}    &\textbf{64.2}         &{60.0}         &\textbf{47.0}       &{61.9}         &\textbf{58.5}         &{49.2}        &{71.5}   &{65.0}&{62.1}        \\
DSI-\FGBMK (ours) &55.1    &{55.4}   &{35.9}    &62.5        &\textbf{60.3}         &46.4       &\textbf{64.4}         &58.4         &\textbf{50.3}        &\textbf{71.6}         &\textbf{65.4}         &\textbf{62.2}        \\\hline
DTSI-FGSM   &49.9     &52.5     &35.2     &52.4         &53.2         &40.9        &56.8         &60.2         &46.2        &73.0         &72.2        &68.4        \\
DTSI-\FGBMN (ours) &{60.3}     &{62.2}     &\textbf{46.4}    &{58.9}         &{59.7}         &{51.8}       &{64.1}         &{66.7}         &{57.2}        &{76.7}         &{74.7}         &{74.5}       \\
DTSI-\FGBMK (ours) &\textbf{60.7}&\textbf{62.9}&{45.6}    &\textbf{62.5}         &\textbf{64.3}        &\textbf{53.3}       &\textbf{68.3}         &\textbf{68.4}         &\textbf{60.2}        &\textbf{80.6}         &\textbf{78.1}         &\textbf{76.9}        \\\hline
MDTSI-FGSM    &\textbf{67.3}     &69.2     &\textbf{54.9}    &\textbf{68.4}         &69.4         &57.8       &67.5         &70.4         &62.1        &78.8         &79.4         &77.5        \\
MDTSI-\FGBMN (ours) &66.7     &69.5     &53.5    &67.6         &68.5         &58.9       &67.4        &70.2         &61.4        &\textbf{79.2}         &80.8         &78.3        \\
MDTSI-\FGBMK (ours) &{67.2}  &\textbf{70.7}  &{54.5}    &\textbf{67.7}         &\textbf{71.7}         &\textbf{59.1}       &\textbf{69.0}         &\textbf{72.5}         &\textbf{62.6}        &{78.8}         &\textbf{80.9}         &\textbf{80.1} \\\hline
\end{tabular}
}
\end{table*}

As shown in Table~\ref{tab:attack_s_NT}, \FGBMN and \FGBMK sustain powerful white-box attack performance since they achieve near 100\% success rate against all white-box models. For black-box attacks, our methods are applicable to most attacks, such as I-FGSM, DI-FGSM, TI-BIM, SI-FGSM. Specifically, by integrating \FGBMN with them, our methods surpass I-FGSM, DI-FGSM, TI-BIM, SIFGSM in the black-box manner by a large margin, \ie, on average, \textbf{9.8\%},\textbf{ 8.6\%}, \textbf{6.8\%}, \textbf{11.4\%} respectively. For MI-\FGBMN, which is based on cumulative gradients, the gradients are so large that most of the pixels will be clipped if applying \FGBMN to them, and therefore no noticeable performance leverage is obtained by MI-\FGBMN compared with MI-FGSM. By integrating \FGBMK with each FGSM-based method, our methods outperform I-FGSM, MI-FGSM, DI-FGSM, TI-BIM, SI-FGSM for the black-box attacks by \textbf{18.7\%}, \textbf{3.3\%}, \textbf{14.1\%}, \textbf{20.3\%}, \textbf{20.7\%}. 

\begin{table}[t]
\caption{Success rates on normally trained models of targeted adversarial attacks compare to different methods against an ensemble of white-box models and a hold-out black-box target model. Where “-” indicates hold-out model. \enquote{*} denotes white-box results while others are black-box results. 
}
\label{tab:targeted_attack_nt}
\centering
\resizebox{1\columnwidth}{!}{
\begin{tabular}{c|c|cccc}\hline
                  Models & Attacks  & -\inca  & -\incb  & -\res   & -\inre \\ \hline\hline
\multirow{12}{*}{Ensemble}  & I-FGSM      &99.9*   &99.9*    &\textbf{100.0*}   &\textbf{100.0*} \\
                            & I-\FGBMN (ours)    &99.9*   &99.9*    &\textbf{100.0*}   &\textbf{100.0*} \\
                            & I-\FGBMK (ours)    &\textbf{100.0*}   &\textbf{100.0*}    &\textbf{100.0*}   &\textbf{100.0*} \\\cline{2-6}
                            & DI-FGSM     &91.7*   &91.3*    &92.2*   &93.7* \\
                            & DI-\FGBMN (ours)   &95.7*   &96.0*    &94.9*   &97.2* \\
                            & DI-\FGBMK (ours)   &\textbf{99.1*}   &\textbf{99.1*}    &\textbf{98.5*}   &\textbf{99.3*} \\\cline{2-6}
                            & TI-BIM     &99.9*   &99.9*    &\textbf{100.0*}   &\textbf{100.0*} \\
                            & \TIBBIMN  (ours)  &99.9*   &99.9*    &\textbf{100.0*}   &\textbf{100.0*} \\
                            & \TIBBIMK (ours)   &\textbf{100.0*}   &\textbf{100.0*}    &\textbf{100.0*}   &\textbf{100.0*} \\\cline{2-6}
                            & DTI-FGSM    &90.3*   &89.7*    &90.5*   &94.4* \\
                            & DTI-\FGBMN (ours)  &95.3*   &96.1*    &95.6*   &96.3* \\
                            & DTI-\FGBMK (ours)  &\textbf{98.2*}   &\textbf{99.0*}    &\textbf{98.1*}   &\textbf{99.4*} \\ \hline\hline
\multirow{12}{*}{Hold-out}  & I-FGSM      &1.2   &0.6    &0.0   &0.4 \\
                            & I-\FGBMN (ours)    &4.6   &3.9    &0.8   &3.3 \\
                            & I-\FGBMK (ours)    &\textbf{10.5}   &\textbf{5.9}    &\textbf{1.5}   &\textbf{6.5} \\\cline{2-6}
                            & DI-FGSM     &15.3   &14.4    &3.6   &8.6 \\
                            & DI-\FGBMN (ours)   &22.5   &25.7    &8.4   &19.5 \\
                            & DI-\FGBMK (ours)   &\textbf{36.9}   &\textbf{36.2}    &\textbf{13.1}   &\textbf{32.1} \\\cline{2-6}
                            & TI-BIM     &2.2   &1.3    &0.4   &0.7 \\
                            & \TIBBIMN (ours)   &6.4   &5.3    &1.6   &3.7 \\
                            & \TIBBIMK (ours)   &\textbf{13.8}   &\textbf{10.7}    &\textbf{3.8}   &\textbf{8.6} \\\cline{2-6}
                            & DTI-FGSM    &16.4   &14.9    &4.3   &10.3 \\
                            & DTI-\FGBMN (ours)  &26.4   &26.8    &9.3   &21.9 \\
                            & DTI-\FGBMK (ours)  &\textbf{39.5}   &\textbf{41.0}    &\textbf{13.7}   &\textbf{35.8} \\ \hline\hline
\end{tabular}
}
\end{table}

\paragraph{Attack against Defense Models}
Adversarial training technique is an effective way to evade attacks, especially for black-box attacks.
Nevertheless, we still obtain expected results. As placed in Table~\ref{tab:attack_s_EAT}, our \FGBMN and \FGBMK achieve the best performance in most cases, especially for I-\FGBMK, DI-\FGBMN, \TIBBIMK, and SI-\FGBMK, which outperform vanilla methods by \textbf{12.0\%}, \textbf{17.2\%}, \textbf{13.3\%} and \textbf{26.4\%} on average. Results among MI variants are also satisfied, where our MI-\FGBMN obtains \textbf{4.2\%} performance gain on average.

Experiment results demonstrate that in most cases, our fixed scale approach \FGBMN reduces the side effects of $\sign$ operation and improves the transferability of adversarial examples. Our adaptive scale approach \FGBMK relaxes restrictions added by \FGBMN with the noise magnitude, adaptively adjusts scale factor, and further improves the attack performance. It is noticeable that performance gap between our \FGBMN and \FGBMK variants is smaller on EAT than NT. Intuitively, due to decision boundaries between EAT and NT being more different than that between NT, adversarial examples generated on NT are more challenging to transfer to EAT,~\ie, smaller gap. Similarly, differences in decision boundaries between EAT and NT result in different attack performance against them using the same K,~\ie, different trends \wrt $K$ in Figure~\ref{fig:Kselect_NT}~and~\ref{fig:Kselect_EAT}.

\begin{table}[t]
\caption{Success rates on adversarially trained models of targeted adversarial attacks compare to different methods against an ensemble of white-box models and a hold-out black-box model. Where “-” indicates the hold-out network. \enquote{*} denotes the white-box results while others are black-box results. 
}
\label{tab:targeted_attack_eat}
\centering
\resizebox{0.82\columnwidth}{!}{
\begin{tabular}{c|c|c|c}\hline
                  Models & Attacks  & Ensemble  & Hold-out\\ \hline\hline
\multirow{12}{*}{-\incaE}   & DTI-FGSM    &87.8*   &3.6\\
                            & DTI-\FGBMN (ours)  &92.8*   &15.4\\
                            & DTI-\FGBMK (ours)  &\textbf{98.5*}   &\textbf{22.1}\\\cline{2-4}
                            & DTMI-FGSM      &83.4*   &9.2\\
                            & DTMI-\FGBMN (ours)   &89.4*   &15.1\\
                            & DTMI-\FGBMK (ours)    &\textbf{97.6*}   &\textbf{12.7}\\\cline{2-4}
                            & DTPoI-FGSM     &83.1*   &1.9\\
                            & DTPoI-\FGBMN (ours)   &91.0*   &11.5\\
                            & DTPoI-\FGBMK (ours)   &\textbf{97.4*}   &\textbf{13.9}\\\cline{2-4}\hline\hline
\multirow{12}{*}{-\incbE}   & DTI-FGSM    &85.9*   &3.0\\
                            & DTI-\FGBMN (ours)  &92.1*   &14.8\\
                            & DTI-\FGBMK (ours)  &\textbf{98.3*}   &\textbf{19.6}\\\cline{2-4}
                            & DTMI-FGSM      &93.2*   &4.5\\
                            & DTMI-\FGBMN (ours)   &95.6*   &\textbf{8.1}\\
                            & DTMI-\FGBMK (ours)    &\textbf{98.9*}   &4.8\\\cline{2-4}
                            & DTPoI-FGSM     &82.5*   &1.9\\
                            & DTPoI-\FGBMN (ours)   &89.8*   &11.3\\
                            & DTPoI-\FGBMK (ours)   &\textbf{97.6*}   &\textbf{12.7}\\\cline{2-4}\hline\hline
\multirow{12}{*}{-\inreE}   & DTI-FGSM    &86.3*   &2.1\\
                            & DTI-\FGBMN (ours)  &93.4*   &\textbf{9.6}\\
                            & DTI-\FGBMK (ours)  &\textbf{98.5*}   &\textbf{9.6}\\\cline{2-4}
                            & DTMI-FGSM      &82.2*   &\textbf{4.4}\\
                            & DTMI-\FGBMN (ours)   &89.9*   &4.3\\
                            & DTMI-\FGBMK (ours)    &\textbf{96.8*}   &3.7\\\cline{2-4}
                            & DTPoI-FGSM     &82.8*   &0.7\\
                            & DTPoI-\FGBMN (ours)   &90.2*   &7.4\\
                            & DTPoI-\FGBMK (ours)   &\textbf{97.7*}   &\textbf{6.9}\\\cline{2-4}\hline\hline
\end{tabular}
}
\end{table}


\paragraph{Effectiveness on Combination of Methods}
To further verify the effectiveness of our methods, we also report the results on different combinations of methods: DTI-FGSM, DSI-FGSM, DTSI-FGSM, MDTSI-FGSM, and their \FGBMN, \FGBMK variants. 
Specifically, we firstly attack given white-box models using different methods and then transfer resultant adversarial examples to black-box models, including NT and EAT. 
As demonstrated in Table.~\ref{tab:attack_c_NT} and~\ref{tab:attack_c_EAT}, our \FGBMN and \FGBMK substantially improve the transferability of black-box attacks without sacrificing the effectiveness of white-box attacks. In most cases, noticeable black-box attack performance gain is achieved by just applying our \FGBMN to original methods, with an average of \textbf{4.1\%}, \textbf{9.9\%} and a maximum of \textbf{11.8\%}, \textbf{24.9\%} for NT and EAT, respectively. The most powerful attacks come from \FGBMK variants. For black-box NT, our \FGBMK variants outperform state-of-the-art attacks by \textbf{11.1\%} on average and \textbf{24.6\%} at most. For EAT, \FGBMK variants achieve comparable results of \FGBMN variants, which outperforms state-of-the-art attacks by \textbf{24.8\%} at most and \textbf{11.3\%} on average. Significant performance improvement on the combined version of attack methods confirms the universality and effectiveness of our approach.

\subsubsection{Performance of targeted attacks}
To verify the performance of proposed \FGBMN  and \FGBMK, following setup of~\cite{mifgsm},~\ie, ensembling models by fusing their logit activations, here we present results of targeted attacks.


\paragraph{Attack against Normally Trained Models}
In Table.~\ref{tab:targeted_attack_nt}, we present success rates of targeted attacks against NT. Adversarial examples are crafted on an ensemble of three models (ensemble model) using different methods, and transferred to remaining one model (hold-out model). Percentage of adversarial examples misclassified as target labels by ensemble and hold-out models represents success rate of targeted white-box and black-box attacks, respectively.
As demonstrated in the table, attack performance of original methods can be improved in all cases by integrating our \FGBMN and \FGBMK with them. Under white-box setting, for example, our DI-\FGBMN, DI-\FGBMK surpass DI-FGSM  by \textbf{3.7\%} and \textbf{6.8\%} (on average), respectively. Under black-box setting, our methods can significantly enhance transferability of adversarial examples regardless of original FGSM-based methods or ensemble models. In particular, when using I-\FGBMK, DI-\FGBMK, \TIBBIMK and DTI-\FGBMK, to transfer to \inca, targeted attack success rates have been increased by \textbf{9.3\%}, \textbf{21.6\%}, \textbf{11.6\%}, and \textbf{23.1\%} than vanilla methods.
Intuitively, a possible reason is that bias of direction between gradients and perturbations caused by sign operations also affects results of targeted attack. Our \FGBMK and \FGBMN mitigate this bias and improve success rate of targeted attacks.

\begin{figure}[!htbp]
    \begin{center}
       \includegraphics[width=1\columnwidth]{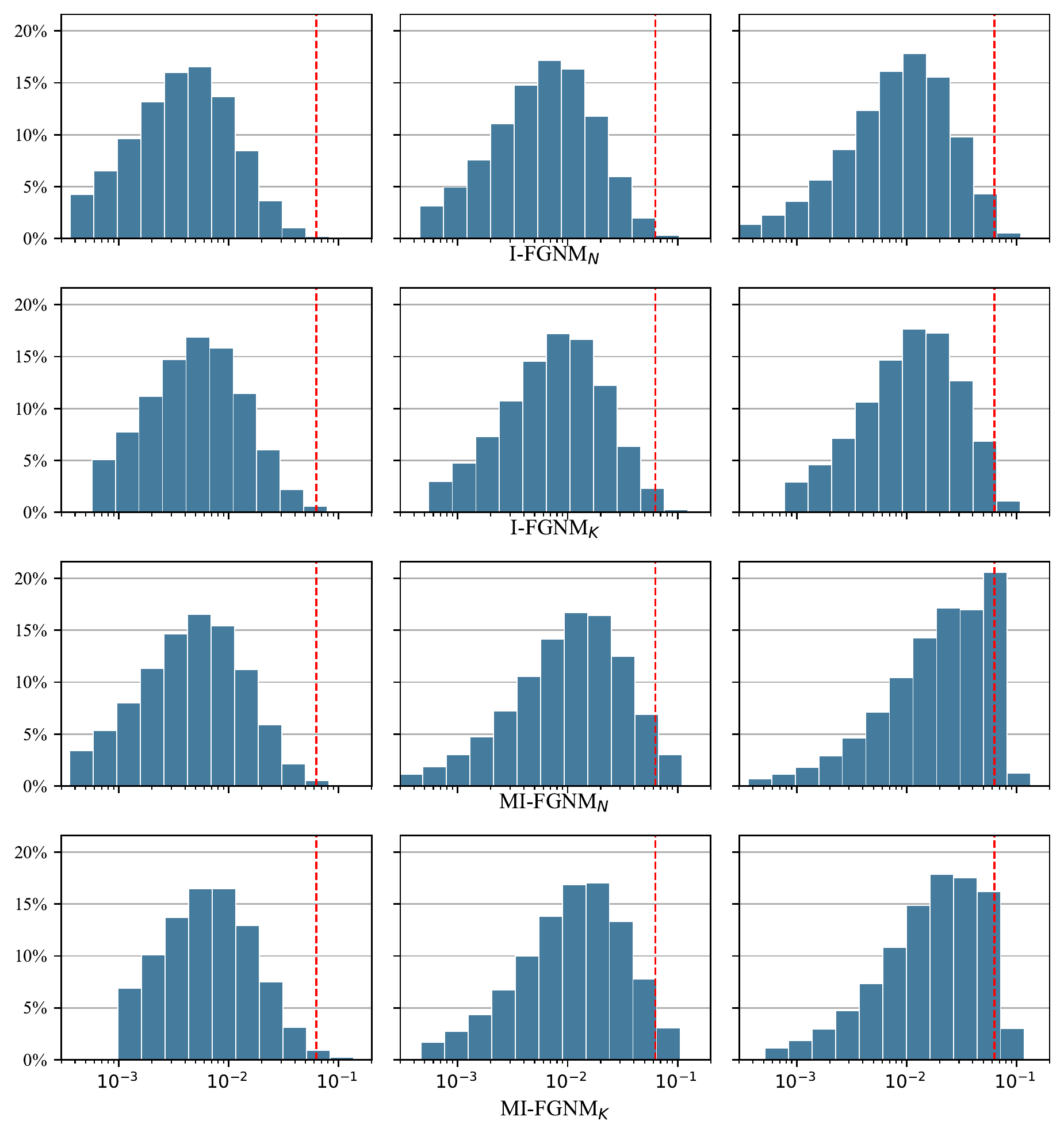}
    \end{center}
   \caption{Histogram of absolute perturbations before clipping at iteration $2$, $5$, $10$ with I-\FGBMN, I-\FGBMK, MI-\FGBMN, and MI-\FGBMK, noises on the right side of the red dashed lines will be clipped. Compared to I-methods, MI-methods produces more perturbations to be clipped.}
    \label{fig:gradientHist}
    \vspace{-0.5cm}
\end{figure}

\paragraph{Attack against Defense Models}
Applying targeted attacks on EAT is more difficult than NT. For the sake of comprehensiveness, we perform targeted attacks using different methods against EAT. Specifically, we firstly attack an ensemble of five models (three NT and two EAT). Then test targeted black-box, white-box attack results on remaining one EAT model (hold-out model) and ensemble model, respectively. Results are presented in Table.~\ref{tab:targeted_attack_eat}, it can be seen that targeted adversarial examples have worse transferability on EAT compared to NT (see Table.~\ref{tab:targeted_attack_nt}). DTI-FGSM, DTMI-FGSM, DTPoI-FGSM demonstrate their effectiveness due to use of diverse inputs, metric learning adopting Poincare distance, and translation-invariant attacks. However, the use of $\sign$ operation limits their attack performance. Our approach \FGBMN and \FGBMK mitigates side-effect of $\sign$ operation through adopting scale factor and further improves success rate of targeted attacks in most cases. Under white-box setting, our \FGBMN, \FGBMK surpassing FGSM-based methods by \textbf{6.3\%}, \textbf{12.7\%} on average and \textbf{7.9\%}, \textbf{15.1\%} at most, respectively. Under black-box setting, it is \textbf{7.4\%}, \textbf{8.3\%} and \textbf{11.8\%}, \textbf{18.5\%}.

\begin{table*}[!htbp]
\centering
\caption{Quantitative comparisons of average magnitude $\norm{\cdot}$ and precision $\cos{\innerprod{\cdot, \vect g}}$ of noises. Given \inca, \incb, \res and \inre as white-box models, adversarial examples are crafted by different methods including 1) FGSM-based,~\eg, IFGSM,~\etc, 2) \FGBMN variants,~\eg, I-\FGBMN,~\etc, 3) \FGBMK variants,~\eg, I-\FGBMK,~\etc, where K is set to 12000.} 
\label{tab:mag and cos}
\resizebox{0.68\textwidth}{!}{
\begin{tabular}{c|cc|cc|cc|cc}\hline
           & \multicolumn{2}{c|}{\inca} & \multicolumn{2}{c|}{\incb} & \multicolumn{2}{c|}{\res} & \multicolumn{2}{c}{\inre} \\
Attacks    &$\norm{\cdot}$  &$\cos{\innerprod{\cdot, \vect g}}$  &$\norm{\cdot}$  &$\cos{\innerprod{\cdot, \vect g}}$       &$\norm{\cdot}$       &$\cos{\innerprod{\cdot, \vect g}}$     &$\norm{\cdot}$       &$\cos{\innerprod{\cdot, \vect g}}$      \\\hline\hline
I-FGSM       &20.21     &0.63     &20.77        &0.61        &21.12         &0.59       &21.12        &0.60        \\
I-\FGBMN (ours)     &\textbf{18.51}     &\textbf{0.98}     &\textbf{18.86}        &\textbf{0.98}        &\textbf{18.52}         &\textbf{0.99}       &\textbf{18.92}        &\textbf{0.98}        \\
I-\FGBMK (ours)     &34.94     &0.92     &36.60        &0.91        &36.48         &0.92       &37.00        &0.90        \\\hline
MI-FGSM      &46.10     &0.64     &47.20        &0.64        &46.90         &0.65       &47.38        &0.63        \\
MI-\FGBMN (ours)    &\textbf{35.91}     &\textbf{0.82}     &\textbf{36.11}        &\textbf{0.81}        &\textbf{35.17}         &\textbf{0.83}       &\textbf{35.69}        &\textbf{0.80}        \\
MI-\FGBMK (ours)    &49.47     &0.60     &50.48        &0.59        &50.39         &0.59       &50.45        &0.58        \\\hline
DI-FGSM      &20.61     &0.60     &21.18        &0.59        &21.40         &0.58       &21.42        &0.58        \\
DI-\FGBMN (ours)    &\textbf{18.76}     &\textbf{0.98}     &\textbf{19.03}        &\textbf{0.97}        &\textbf{18.86}         &\textbf{0.99}       &\textbf{18.98}        &\textbf{0.97}        \\
DI-\FGBMK (ours)    &36.56     &0.90     &37.47        &0.89        &37.39         &0.89       &37.62        &0.87        \\\hline
TI-BIM      &21.33     &0.67     &21.89        &0.64        &23.06         &0.61       &23.07        &0.63        \\
\TIBBIMN (ours)     &\textbf{19.46}     &\textbf{0.99}     &\textbf{19.96}        &\textbf{0.98}        &\textbf{19.86}         &\textbf{0.98}       &\textbf{20.23}        &\textbf{0.98}        \\
\TIBBIMK (ours)    &34.61     &0.94     &35.99        &0.93        &36.85         &0.90       &36.70        &0.91        \\\hline
SI-FGSM      &22.48     &0.65     &23.00        &0.64        &22.80         &0.57       &23.46        &0.62        \\
SI-\FGBMN (ours)    &\textbf{19.99}     &\textbf{0.99}     &\textbf{20.15}        &\textbf{0.99}        &\textbf{19.57}         &\textbf{0.97}       &\textbf{20.25}        &\textbf{0.98}        \\
SI-\FGBMK (ours)    &35.68     &0.92     &36.57        &0.93        &36.55         &0.91       &37.01        &0.89        \\\hline
DTI-FGSM     &21.62     &0.65     &21.89        &0.64        &22.75         &0.62       &23.06        &0.63        \\
DTI-\FGBMN (ours)   &\textbf{19.90}     &\textbf{0.98}     &\textbf{19.96}        &\textbf{0.98}        &\textbf{20.39}         &\textbf{0.98}       &\textbf{20.22}        &\textbf{0.98}        \\
DTI-\FGBMK (ours)   &35.65     &0.93     &36.99        &0.93        &37.67         &0.90       &36.72        &0.91        \\\hline
DSI-FGSM     &23.10     &0.64     &23.57        &0.63        &23.44         &0.62       &24.03        &0.59        \\
DSI-\FGBMN (ours)   &\textbf{20.55}     &\textbf{0.98}     &\textbf{20.52}        &\textbf{0.98}        &\textbf{20.27}         &\textbf{0.97}       &\textbf{20.62}        &\textbf{0.97}        \\
DSI-\FGBMK (ours)   &36.17     &0.92     &36.90        &0.89        &36.79         &0.86       &37.33        &0.88        \\\hline
DTSI-FGSM    &25.71     &0.67     &26.00        &0.64        &26.60         &0.66       &28.19        &0.62        \\
DTSI-\FGBMN (ours)  &\textbf{23.23}     &\textbf{0.98}     &\textbf{22.99}        &\textbf{0.97}        &\textbf{23.42}         &\textbf{0.97}       &\textbf{24.16}        &\textbf{0.96}        \\
DTSI-\FGBMK (ours)  &37.09     &0.92     &37.61        &0.90        &38.38         &0.85       &38.74        &0.87        \\\hline
MDTSI-FGSM   &41.48     &0.67     &41.71        &0.62        &42.25         &0.66       &43.87        &0.62        \\
MDTSI-\FGBMN (ours) &\textbf{34.67}     &\textbf{0.88}     &\textbf{34.25}        &\textbf{0.85}        &\textbf{34.22}         &\textbf{0.86}       &\textbf{33.93}        &\textbf{0.82}        \\
MDTSI-\FGBMK (ours) &47.56     &0.71     &47.75        &0.67        &48.03         &0.67       &48.21        &0.62        \\\hline
\end{tabular}
}
\end{table*}

\begin{figure*}[!htbp]
    \begin{center}
       \includegraphics[width=\textwidth]{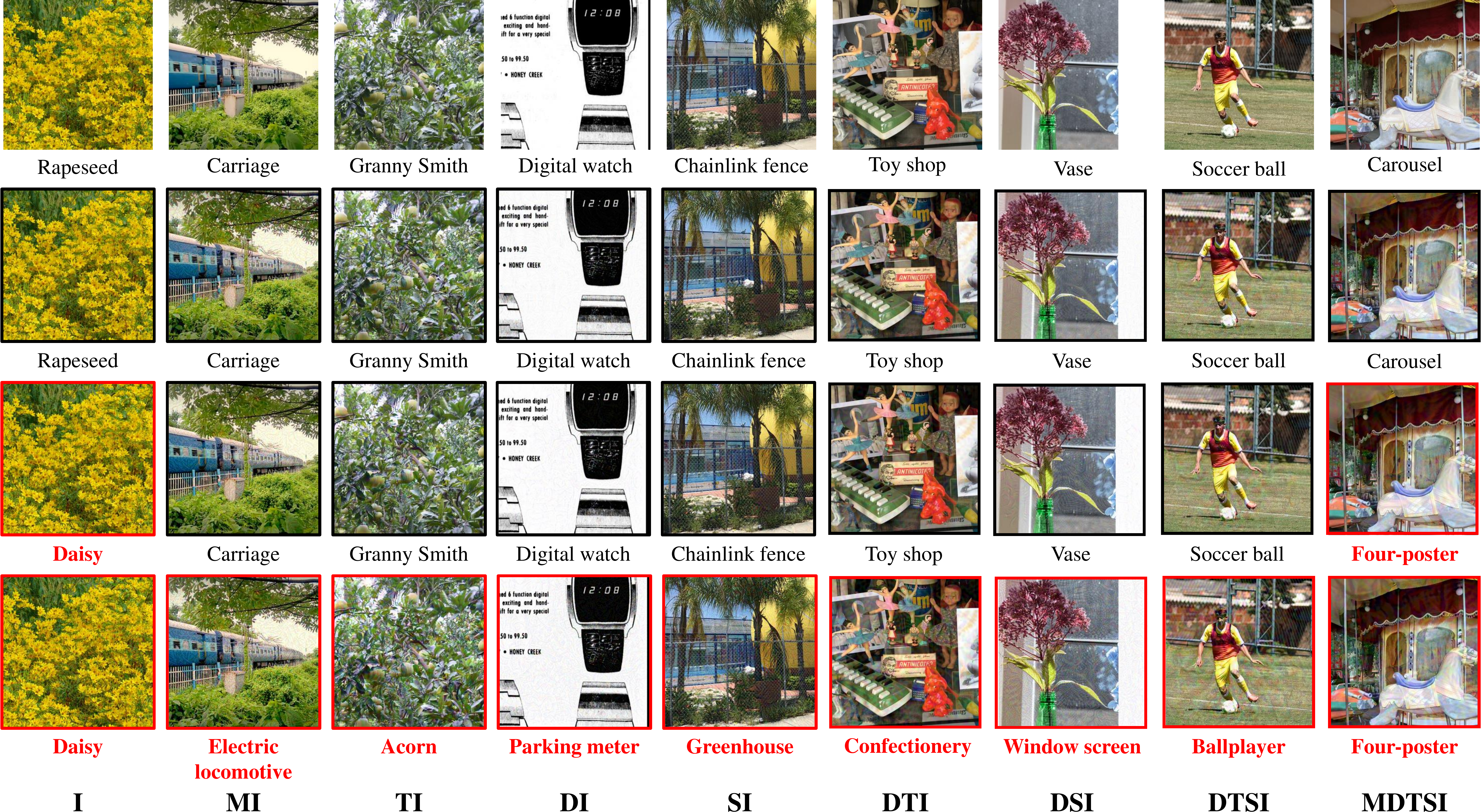}
    \end{center}
   \caption{Visualization of adversarial examples. From top to bottom row-by-row: 1) raw images, 2) methods using $\sign$ operation, 3) methods with our fixed scale, \eg, I-\FGBMN \etc, and 4) methods with our adaptive scale, \eg, \TIBBIMK \etc, where $K$ is set to $12,000$. From left to right are the basic algorithms adopted. We choose images generated by \inca white-box and tested them on \incb. The predictions are placed below the images, and misclassified are highlighted in red.}
    \label{fig:visualization}
\end{figure*}

\subsection{Analysis}
\subsubsection{Magnitude and precision of noise}
For verifying that our method reduces side effects of $\sign$ operation, we report magnitude ($\ell_2$-norm $\norm{\cdot}$) and precision ($\cos{\innerprod{\cdot, \vect{g}}}$) of adversarial noises in Table.~\ref{tab:mag and cos}. Given untargeted adversarial examples generated via NT using different methods, magnitude and precision of noises are averaged among the whole iteration.
It can be observed that our \FGBMN variants generate smaller and more accurate noises compared to original methods,~\ie, closer to gradient direction. For example, noises generated by I-\FGBMN, MI-\FGBMN is on average \textbf{2.1\%}, \textbf{11.2\%} smaller than corresponding FGSM-based methods. 
It is worth noting that cos value of our fixed scale approach \FGBMN does not reach 1.0 due to clip operation, but it is still higher than original methods I-FGSM, MI-FGSM by \textbf{0.4} and \textbf{0.2} (on average), respectively. Compared with \FGBMN, our \FGBMK methods generate larger noise magnitudes and smaller cos values because of relaxed constraints on noise magnitudes. Nevertheless, cos value of \FGBMK method is still higher than that of original FGSM-based method by \textbf{0.2} on average and \textbf{0.3} at most. Obviously, both \FGBMN and \FGBMK alleviate side effects of sign operation and reduce gaps between real gradients and actual noises.

\subsubsection{Analysis of MI-FGNM}
Experimental results show that the performance gain on MI-\FGBMN is not significant compared with other methods. A potential reason is that MI-FGSM applies momentum term to the perturbations generation. The accumulated noises lead to a larger clipped ratio than other methods and hinder the performance of MI-\FGBMN. Further experiments are conducted to verify this. Given Inc-v3 as the white-box model, we report the magnitude of noises before clipping over the whole dataset at iteration 2, 5, and 10 using various methods, including I-\FGBMN, I-\FGBMK, MI-\FGBMN, MI-\FGBMK. As demonstrated in Figure~\ref{fig:gradientHist}, the clipped ratio of I-\FGBMN at iteration 2, 5, and 10 is 0.07\%, 0.03\% and 0.8\% respectively. For I-\FGBMK, it is 0.2\%, 0.8\%, and 1.5\%. For MI-\FGBMN, it is 0.2\%, 3.7\%, 15\%. For MI-\FGBMK, it is 0.6\%, 3.5\%, 7.6\%. In brief, no more than $0.8\%$ of perturbations in I-\FGBMN will be clipped, but for MI-\FGBMN, the percentage increases to $15.0\%$. The larger clipped ratio distorts the direction of perturbations and decreases the magnitude of the final noise, which further hinders performance of MI-\FGBMN. Unlike MI-\FGBMN, MI-\FGBMK considers the trade-off between magnitude and bias to increase performance.



\subsubsection{Visualization}
Visualizations of adversarial examples are given to give qualitative analysis and confirm the effectiveness of generated perturbations. Specifically, we randomly choose several successful adversarial images for each approach, \ie, I-FGSM, MI-FGSM, DI-FGSM, TI-BIM, SI-FGSM, DTI-FGSM, DSI-FGSM, DTSI-FGSM, and MDTSI-FGSM and plot the results in Figure~\ref{fig:visualization}. In this figure, the first row is the raw image while the remaining rows are adversarial examples crafted by original methods and ours {\FGBMN} and {\FGBMK} variants, respectively. As shown in the figure, noises are difficult to distinguish between original methods and ours visually. Notably, our {\FGBMN} variants produce nearly the same adversarial examples with originals, and {\FGBMK} variant slightly increases the noise level. The visualization indicates that our methods can easily boost performance without adding significant noises.

\section{Conclusion}
\label{Sec.Conclusion}
Based on the Taylor expansion, we conduct a comprehensive analysis of gradient-based methods' $\sign$ operation and reveal the gap between real gradients and resultant noises, leading to biased and inefficient attacks. A novel routine named Fast Gradient Non-sign Methods is proposed to correct this gap. Notably, FGNM is a general routine that seamlessly replaces the conventional $sign$ operation in gradient-based attacks with negligible extra computational cost. To confirm the effectiveness of our FGNM, extensive experiments under non-targeted and targeted $\ell_\infty$ settings are conducted. Significant performance gains are observed compared with state-of-the-art methods. For untargeted black-box attacks, ours outperform them by \textbf{27.5\%} at most and \textbf{9.5\%} on average. For targeted attacks against defense models, it is \textbf{15.1\%} and \textbf{12.7\%}.

\newpage
\bibliographystyle{IEEEtran}
\bibliography{bare_jrnl_new_sample4}
\vfill

\end{document}